\documentclass{article}

\usepackage{arxiv}

\usepackage[utf8]{inputenc} 
\usepackage[T1]{fontenc}    
\usepackage{hyperref}       
\usepackage{url}            
\usepackage{booktabs}       
\usepackage{amsfonts}       
\usepackage{nicefrac}       
\usepackage{microtype}      
\usepackage{lipsum}		
\usepackage{graphicx}
\usepackage{natbib}
\usepackage{doi}
\usepackage{natbib}
\usepackage{amsmath,amssymb,graphicx}
\usepackage{algorithm}
\usepackage{algpseudocode}
\DeclareMathOperator*{\argmax}{arg\,max}

\usepackage{comment}
\usepackage{tikz}
\usepackage{algpseudocode}
\usetikzlibrary{positioning,arrows.meta, calc, fit, patterns}
\usetikzlibrary{decorations.pathreplacing}

\usepackage{amsthm}
\theoremstyle{definition}
\newtheorem{definition}{Definition}
\newtheorem{assumption}{Assumption}
\newtheorem{proposition}{Proposition}

\title{Budgeted Subset Selection with a Transferable Learned Value Model}

\author{Jade Lejeune Herman \\
	Mechatronics, Biostatistics and Sensors (MeBioS) \\
	KU Leuven \\
	Kasteelpark Arenberg 30, B-3001 Leuven (Heverlee) \\
	\texttt{jade.lejeuneherman@kuleuven.be}
	\And
	Johan A.K. Suykens \\
	ESAT-Stadius \\
	KU Leuven \\
	Kasteelpark Arenberg 10, B-3001 Leuven (Heverlee) \\
	\texttt{johan.suykens@kuleuven.be}
	\And
	Peter Goos \\
	Mechatronics, Biostatistics and Sensors (MeBioS) \\
	KU Leuven \\
	Kasteelpark Arenberg 30, B-3001 Leuven (Heverlee) \\
	\texttt{peter.goos@kuleuven.be}
}

\renewcommand{\shorttitle}{\textit{arXiv} Template}

\begin{document}
\maketitle
\begin{abstract}
Subset selection is a fundamental task in machine learning, statistics and combinatorial optimisation, with applications in active learning, band selection, input selection and experimental design. The utility of a subset depends on information revealed only after the selection, such as labels or responses, and on interactions among selected candidates. Existing approaches combine value estimation and redundancy control in application-specific ways, which limits their use in other domains. This paper introduces Subset Acquisition with Learned Value Estimates and Search (SALVES), a framework that decomposes subset selection into two components. First, a value model is trained by supervised learning on tasks where the hidden information is known. On a new task, it predicts the utility each candidate would add to the current selection. Second, a selection step chooses the subset from these predictions, typically by simulated annealing with a diversity term penalising redundancy. The value model transfers to new classes, pretrained networks, datasets and sensors, and from synthetic to real data. On the four applications, SALVES matches or outperforms application-specific baselines on average. It requires substantially fewer bands for equal accuracy and remains reliable in input selection where individual baselines fail. 
\end{abstract}


\newpage
{\color{black}
\section{Introduction}
Subset selection is a fundamental task in machine learning, statistics and combinatorial optimisation. It arises in a variety of settings, such as selecting images for annotation, identifying informative spectral bands, choosing input variables for a predictive model or determining which modifications of an experimental design to evaluate. Despite their diversity, these applications share a common structure. They select a small number of candidates from a large set, and the utility of the selection depends on the subset as a whole rather than on its candidates individually.

Formally, consider $N$ candidates indexed by $[N]=\{1,\dots,N\}$, and let $U:2^{[N]}\rightarrow\mathbb{R}$ be a set function, called the utility, that assigns a value to each subset of candidates. Given a selection budget $n<N$, the objective is to find
\begin{equation}
L^\star\in\arg\max_{L\subseteq[N],\,|L|\le n}U(L).
\label{eq:intro-problem}
\end{equation}
The utility captures the quality of a selected subset and depends on the application. For instance, in input selection, the candidates are the available input variables, and $U(L)$ may measure the predictive performance of a model trained using the variables in $L$.

These problems share three fundamental challenges. First, the number of possible subsets, $\binom{N}{n}$, is typically too large for exhaustive enumeration. Second, the quality of a candidate subset depends on information that is hidden when the selection is made, such as the labels of the candidates. It can only be assessed after a costly procedure, such as collecting labels or training a model on the selected candidates. Third, the utility of a subset is not simply the sum of the contributions of its candidates. Candidates may interact, so that selecting several redundant candidates adds little over selecting one of them.

These challenges have been studied extensively in the literature, as discussed in Section~\ref{sec:related}. However, existing methods have largely been developed for specific problem settings, with objectives and constraints tailored to their respective applications. In active learning, for example, candidates are typically ranked by an acquisition score \citep{ref51}, often combined with clustering or core-set constructions to avoid redundant queries \citep{Ref8, Ref11, Ref12, Ref9}. In input and band selection, filter methods rank candidates by their association with the target \citep{Ref18}, while redundancy is controlled through correlation or mutual-information penalties \citep{Ref16} or through a clustering of the candidates \citep{Ref14}.

More generic approaches, such as greedy maximisation of submodular functions \citep{ref52, ref32} or mixed-integer optimisation \citep{ref64, ref65}, can be applied across different domains and provide guarantees under suitable assumptions. However, their applicability depends on the objective function satisfying specific structural properties, and often on the ability to evaluate it efficiently. These conditions may not hold in many practical subset selection problems.

Despite their different origins, these methods share a common structure. Each one, explicitly or implicitly, estimates the value of candidates from the information available at selection time and constructs a subset that balances high estimated value against redundancy. However, methods developed in one domain are seldom applied in another, because these two components are designed jointly for a specific task, and improvements to one component cannot easily be combined with alternatives for the other.

The first of these two components can also be learned rather than designed by hand. In active learning, the benefit of labelling a candidate has been learned on tasks whose labels are available and then applied to new tasks \citep{ref22, ref23, ref24, ref25}, and in combinatorial optimisation, a learned score guides a classical search \citep{ref26, ref27, ref28}. However, each of these lines of work remains specific to one family of problems.

Building on these observations, this paper proposes SALVES (Subset Acquisition with Learned Value Estimates and Search), a general framework that decomposes subset selection into two explicit components. The first is a value model, which predicts the gain of each candidate, that is, the utility it adds to the subset already selected, from a state that summarises the information available at selection time. The value model is learned on training tasks for which the utility can be evaluated and is then applied to new tasks. This addresses the second challenge: since the utility cannot be evaluated during selection, the predicted gains are used in its place. The second is a diversity term, which penalises redundancy among the selected candidates and addresses the third challenge. In most applications, a search by simulated annealing \citep{Ref1} then selects a subset that combines high predicted gains with diversity. Because the search never enumerates subsets, the framework also avoids the combinatorial explosion of the first challenge.

SALVES is evaluated on four selection problems: batch active learning, band selection in hyperspectral imaging, input selection and experimental design. The contributions of this paper are the following.
\begin{itemize}
\item A formulation of budgeted selection under hidden information over a distribution of tasks, together with a sequence of questions that decides whether a new problem belongs to this class and in which form SALVES applies to it.
\item The SALVES framework, which learns on training tasks a gain that cannot be computed on the new task, describes the candidates by signals that keep their meaning from one task to another and combines the predicted values with diversity in the selection of a batch. Classical selection rules are special cases of the framework in which the value is fixed by hand.
\item A result showing that, for a classifier trained by minimising the cross-entropy, the expected gain of a label cannot be estimated from the classifier alone, which explains why the value has to be learned on tasks where the hidden information is known.
\item An evaluation of a single framework on four problems from different communities, including transfer to new classes, to new pretrained networks, to new datasets, to a new sensor and from synthetic to real data, on pools of very many candidates.
\item An analysis of which components account for the results, which attributes the gain in active learning and band selection to the value model and to diversity rather than to the choice of the search algorithm.
\end{itemize}

The remainder of the paper is organised as follows. Section~\ref{sec:related} reviews related work. Section~\ref{sec:problem} formalises the selection problem and states the assumptions underlying the framework. Section~\ref{sec:salve} describes SALVES, followed by Section~\ref{sec:scope}, which discusses its scope and positions it relative to existing approaches. Section~\ref{sec:applications} introduces the four applications, and Section~\ref{sec:protocol} details the experimental protocol. Section~\ref{sec:ablations} presents the ablation studies. Section~\ref{sec:res-setup} summarises the experimental setup. Sections~\ref{sec:res-al} to~\ref{sec:res-design} report the results for the individual applications, while Section~\ref{sec:res-analyses} examines the contribution of the different components of SALVES to these results.
\section{Related work}
\label{sec:related}
Existing subset selection methods are typically designed by hand and tailored to a specific application. In batch active learning, candidates are scored by the uncertainty of the classifier \citep{ref51, Ref3}, by how well they represent or cover the pool \citep{Ref8, Ref9, Ref10} or by a combination of uncertainty and diversity \citep{Ref11, Ref12}. In band selection and input selection, candidates are ranked by their association with the target \citep{Ref18} or by the weights of a fitted model \citep{Ref15}, and redundancy is handled by a penalty \citep{Ref16} or by a clustering of the candidates \citep{Ref14}. In experimental design, designs are constructed by exchange algorithms \citep{ref54}, by simulated annealing with modifications proposed at random \citep{ref55} or by genetic algorithms \citep{ref69}. Several studies report that no rule is the strongest across tasks and that the ranking of the rules changes with the budget and the dataset \citep{Ref9, Ref18, ref19b, ref20}.

Several lines of work learn components of a selection rule instead of designing them by hand, each addressing an aspect of the problem considered by SALVES. In active learning, the value of labelling a candidate has been regressed on tasks where the labels are known and applied to new tasks, and oracles that use the labels have been imitated \citep{ref22, ref23, ref24, ref25}. These rules select one candidate at a time or take the highest scores without an explicit diversity term, and they have been evaluated on much smaller problems than those considered here. In combinatorial optimisation, a score per variable or node is learned from features that are cheap to compute and inserted in a classical search, for instance to decide where to branch or which nodes to keep \citep{ref26, ref27, ref28}. In these methods, the quantity that the score imitates can still be computed on the new instance, only at a high cost, so the benefit of learning is speed.

Learned rules that transfer to new tasks also exist in band selection and in input selection. In band selection, graph networks trained on several hyperspectral scenes, by reinforcement learning \citep{ref56} or with existing selection methods as teachers \citep{ref57}, have been applied without further training to new scenes, including scenes recorded by other sensors. In input selection, a network trained on synthetic regression tasks in which the informative variables are known predicts, for a new dataset, the probability that each variable is informative \citep{ref58}. SALVES shares the principle of these methods and differs in its inputs. Its value model reads signals that keep their meaning from one task to another, so that the same small model applies to tasks of very different sizes.

Selection rules have also been learned by reinforcement learning, with the utility as the reward, both in active learning \citep{ref36} and for routing problems such as the travelling salesman problem, where a tour is constructed one city at a time \citep{ref59, ref37, Ref35}. Reinforcement learning is well suited to routing because the contribution of a decision depends on the decisions that follow. In the problems considered here, the gain of each candidate can be computed on the training tasks, so the value model is trained by supervised learning, and no credit has to be assigned over a sequence of decisions.

In predict-then-optimise, the unknown coefficients of an objective are predicted from features and the objective is then optimised, but nothing is revealed during the optimisation and the instances are assumed to be independent \citep{ref29, ref30, ref31}. In adaptive selection, the candidate with the highest expected gain given the revealed information is chosen, but this expectation is computed from a known model of a single task \citep{ref32, ref33}. Bayesian experimental design follows the same principle, with the expected information gain as the criterion \citep{ref62}. Policies that choose the next experiment have also been trained in advance on simulated experiments, which again requires a known model \citep{ref63}. In data valuation, values are assigned to training examples \citep{Ref2, ref34}. These values are usually recomputed on each dataset, although functions that weight or select labelled examples have been learned on one dataset and applied to others \citep{ref60, ref61}. SALVES brings these ideas together in a single framework. Its value model learns, on training tasks, a gain that cannot be computed on the new task, because the gain depends on hidden information, such as the labels of the candidates, that is not available there. The complete batch is then selected from the predicted gains, typically in combination with a diversity term.}
\section{Methodology}
\subsection{Problem and scope}
\label{sec:problem}
\textcolor{black}{The problems considered in this paper consist in selecting a small subset from a large set of candidates, when the quality of the selection can only be assessed after a costly procedure. This subsection formulates these problems within a common notation and specifies the assumptions under which the proposed framework applies. Compared with the statement given in the introduction, the formulation adds two notions. The problem is defined for a task, so that several tasks can be related to one another, and the selection proceeds in rounds, so that the information revealed during the selection can be used.}

A task, denoted by $\tau$, is one instance of the selection problem. It consists of $N$ candidates, indexed by $i\in[N]$. Candidate $i$ is described by a vector $x_i\in\mathbb{R}^d$, where $d$ is the dimension of the description, and $\mathcal{X}=(x_1,\dots,x_N)$ collects the descriptions of all candidates. A subset of candidates is represented by a set of indices $L\subseteq[N]$. The goal is to select a subset of $n$ candidates, where the budget $n$ is smaller than $N$.

The quality of a selected subset $L$ is measured by its utility $U_\tau(L)\in\mathbb{R}$, which is returned by an evaluation procedure fixed before the selection takes place. For example, the evaluation procedure may train a model on the selected candidates and measure its accuracy on held-out data. The objective is to maximise the utility within the budget $n$. Reaching this objective is difficult because utility evaluations are costly and because the utility depends on information that is not yet observed at the moment the candidates are chosen. This hidden information is denoted by $\omega_\tau$, such that
\begin{equation}
U_\tau(L)=u\big(L,\omega_\tau\big),
\label{eq:utility}
\end{equation}
where $u$ denotes the evaluation procedure, which also uses the descriptions $\mathcal{X}$.

The $n$ candidates are not necessarily selected all at once. The selection is divided into successive steps, called rounds, and the hidden information of a candidate becomes known only at the end of the round in which it is selected. In each round, a group of candidates, called a batch, is selected. The number of rounds is denoted by $T\ge 1$, and the rounds are indexed by $t=1,\dots,T$. The selection may start from candidates whose hidden information is already known. They form the initial set $L_0\subseteq[N]$, which is empty when no such candidate exists. At round $t$, a batch $S_t$ of $B_t$ candidates is selected among those not yet selected, so that $S_t\subseteq[N]\setminus L_{t-1}$, and the selected set becomes $L_t=L_{t-1}\cup S_t$. The batch sizes $B_1,\dots,B_T$ are fixed in advance and satisfy $\sum_{t=1}^{T}B_t=n$. At the end of round $t$, the hidden information concerning the candidates selected so far becomes known. This known part of $\omega_\tau$ is called the revealed information and is denoted by $\omega_\tau(L_t)$. It can be used in the following rounds. With a single round, $T=1$, the entire subset is selected at once. With several rounds, $T>1$, each selection can depend on the information revealed in the earlier rounds. In both cases, the usefulness of a candidate is unknown at the moment it is chosen.

The selection problem can now be stated for one round. Let $\mathcal{S}_t=\{S\subseteq[N]\setminus L_{t-1}:|S|=B_t\}$ denote the set of admissible batches at round $t$, that is, all sets of $B_t$ candidates not yet selected. The best batch at round $t$, denoted by $S_t^\star$, is an admissible batch that gives the highest utility once added to the candidates already selected,
\begin{equation}
S_t^\star\in\arg\max_{S\in\mathcal{S}_t}\;U_\tau\big(L_{t-1}\cup S\big).
\label{eq:problem}
\end{equation}
This problem cannot be solved directly. The number of admissible batches is far too large for exhaustive enumeration, and the utility of each of them depends on hidden information that is not yet revealed. What is known at the start of round $t$ is called the state and is denoted by $s_{t-1}$. It consists of the descriptions of the candidates, the set already selected and its revealed information,
\begin{equation}
s_{t-1}=\big(\mathcal{X},\,L_{t-1},\,\omega_\tau(L_{t-1})\big).
\label{eq:state}
\end{equation}
The goal is to choose, at each round and from the current state alone, a batch $S_t$ whose utility $U_\tau(L_{t-1}\cup S_t)$ is as close as possible to the utility $U_\tau(L_{t-1}\cup S_t^\star)$ of the best batch $S_t^\star$. This choice is made by a selection rule, denoted by $\pi$, a function that maps the current state to an admissible batch, so that $S_t=\pi(s_{t-1})$. The same rule should generalise to new tasks that were not available when it was built. Each round is treated as a separate problem, and its batch is chosen for the utility it adds at that round, as in all the selection rules compared in this paper. \textcolor{black}{This formulation is myopic: it leaves aside a second role of a batch, which is to reveal information that guides the following rounds \citep{ref32}.}

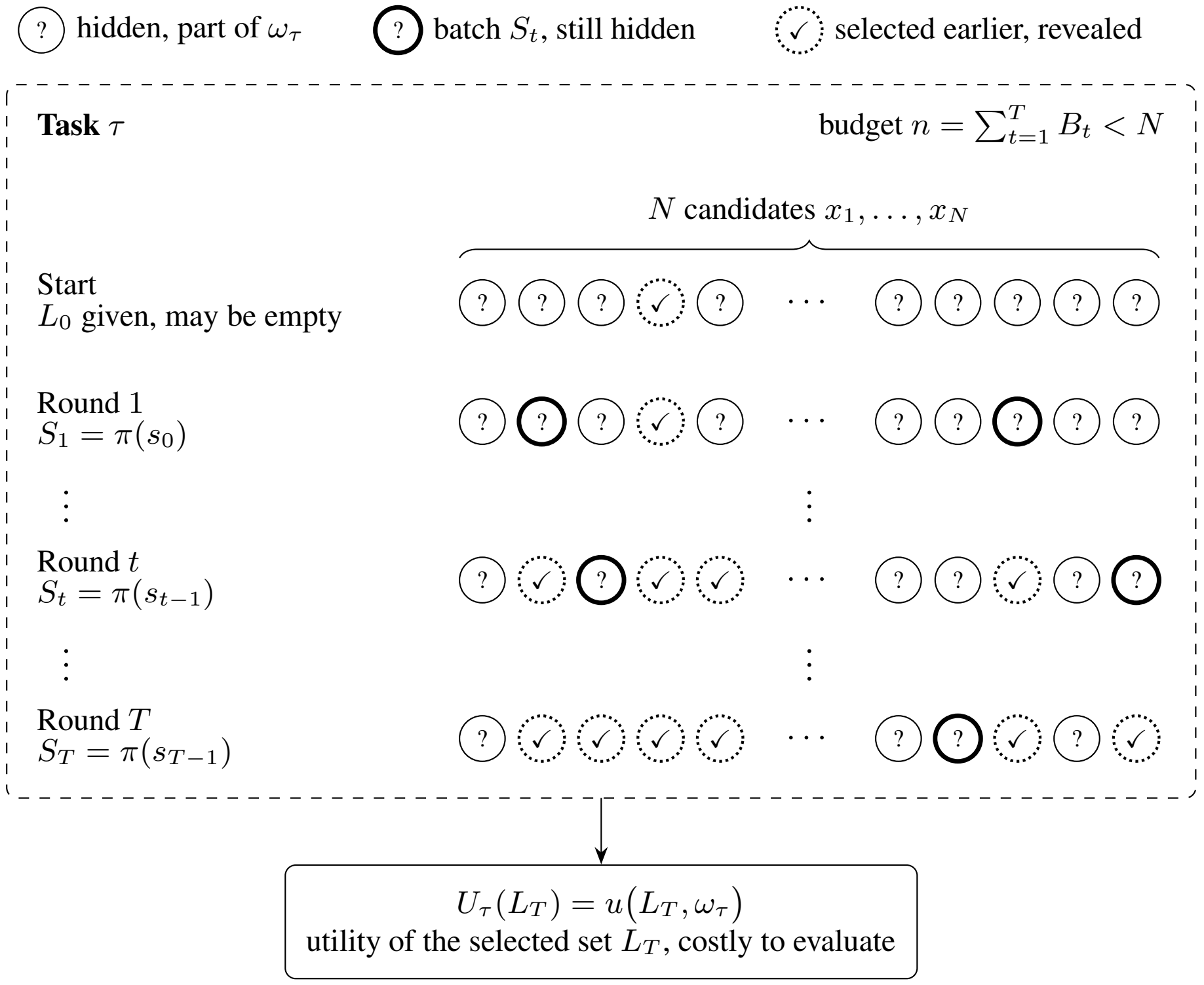
\begin{figure}[t]
    \centering
    \begin{tikzpicture}[
    cand/.style={circle, draw=black, minimum size=4.6mm, inner sep=0pt,
                 font=\scriptsize},
    hid/.style ={cand, fill=white, line width=0.4pt,
                 execute at begin node={?}},
    pick/.style={cand, fill=white, line width=1.3pt,
                 execute at begin node={?}},
    rev/.style ={cand, fill=white, densely dotted, line width=0.8pt,
                 execute at begin node={$\checkmark$}},
    rowlabel/.style={anchor=west, align=left, font=\small, inner sep=0pt},
    legendtext/.style={anchor=west, font=\footnotesize, inner sep=0pt},
  ]

  \newcommand{\candrow}[2]{%
    \foreach \s [count=\k from 0] in {#2} {%
      \pgfmathsetmacro{\x}{0.6*\k + (\k>4)*1.2}%
      \node[\s] at (\x,#1) {};%
    }%
    \node at (3.3,#1) {$\cdots$};%
  }

  \node[hid]  at (-4.45,2.75) {};
  \node[legendtext] at (-4.10,2.75) {hidden, part of $\omega_\tau$};
  \node[pick] at (-0.85,2.75) {};
  \node[legendtext] at (-0.50,2.75) {batch $S_t$, still hidden};
  \node[rev]  at (3.20,2.75) {};
  \node[legendtext] at (3.55,2.75) {selected earlier, revealed};

  \draw[dashed, rounded corners=3pt] (-4.8,2.2) rectangle (7.2,-5.0);
  \node[anchor=west, font=\small\bfseries, inner sep=0pt] at (-4.5,1.8)
       {Task $\tau$};
  \node[anchor=east, font=\footnotesize, inner sep=0pt] at (6.9,1.8)
       {budget $n=\sum_{t=1}^{T}B_t<N$};

  \draw[decorate, decoration={brace, amplitude=5pt}]
       (-0.23,0.45) -- (6.83,0.45)
       node[midway, above=6pt, font=\footnotesize]
       {$N$ candidates $x_1,\dots,x_N$};

  \node[rowlabel] at (-4.5,0)
       {Start\\[-1pt]\footnotesize $L_0$ given, may be empty};
  \candrow{0}{hid,hid,hid,rev,hid,hid,hid,hid,hid,hid}

  \node[rowlabel] at (-4.5,-1.2)
       {Round $1$\\[-1pt]\footnotesize $S_1=\pi(s_0)$};
  \candrow{-1.2}{hid,pick,hid,rev,hid,hid,hid,pick,hid,hid}

  \node at (-4.2,-1.95) {$\vdots$};
  \node at (3.3,-1.95)  {$\vdots$};

  \node[rowlabel] at (-4.5,-2.8)
       {Round $t$\\[-1pt]\footnotesize $S_t=\pi(s_{t-1})$};
  \candrow{-2.8}{hid,rev,pick,rev,rev,hid,hid,rev,hid,pick}

  \node at (-4.2,-3.55) {$\vdots$};
  \node at (3.3,-3.55)  {$\vdots$};

  \node[rowlabel] at (-4.5,-4.4)
       {Round $T$\\[-1pt]\footnotesize $S_T=\pi(s_{T-1})$};
  \candrow{-4.4}{hid,rev,rev,rev,rev,hid,pick,rev,hid,rev}

  \node[draw, rounded corners=3pt, align=center, inner sep=6pt, font=\small]
       (util) at (1.2,-6.25)
       {$U_\tau(L_T)=u\big(L_T,\omega_\tau\big)$\\[1pt]
        \footnotesize utility of the selected set $L_T$, costly to evaluate};
  \draw[-{Stealth[length=5pt]}] (1.2,-5.0) -- (util.north);

\end{tikzpicture}
    \caption{Batch selection over rounds.}
    \label{fig:setting}
\end{figure}

Figure~\ref{fig:setting} illustrates this process. Each row shows the task at one moment of the selection, and each circle is a candidate. At round $t$, the selection rule chooses the batch $S_t$, drawn with thick circles, from the state $s_{t-1}$, at a moment when the hidden information of these candidates is still unknown. At the end of the round, this information is revealed, and in the following rows the batch appears with dotted circles, as part of the revealed information on which the next choices rely. After the last round, the utility of the selected set $L_T$ is evaluated, which is the costly step of the problem.

\textcolor{black}{The usefulness of a candidate is measured by the gain in utility obtained by adding it to the selected set.}

\begin{definition}[\textcolor{black}{Gain of a candidate}]\label{def:gain}
\textcolor{black}{Let $s=(\mathcal{X},L,\omega_\tau(L))$ be a state of task $\tau$ and let $i\in[N]\setminus L$ be a candidate not yet selected. The gain of candidate $i$ in state $s$ is}
\begin{equation}
v_i(s)=U_\tau\big(L\cup\{i\}\big)-U_\tau(L).
\label{eq:gain}
\end{equation}
\end{definition}

\textcolor{black}{The gain depends on the hidden information of candidate $i$, so it cannot be computed when the candidate is chosen. It is defined for a single candidate, and the utility of a batch is in general not the sum of the gains of its candidates, since candidates that carry the same information do not add their gains. The proposed framework relies on three assumptions.}

\begin{assumption}[\textcolor{black}{Training tasks}]\label{ass:tasks}
\textcolor{black}{A set of $M$ tasks $\tau_1,\dots,\tau_M$, called training tasks, is available for which the hidden information is known. On a training task, the utility $U_\tau(L)$ of any subset $L$ can therefore be evaluated.}
\end{assumption}

\begin{assumption}[\textcolor{black}{Predictable gain}]\label{ass:predict}
\textcolor{black}{The gain $v_i(s)$ can be predicted, at least partly, from the state $s$. Although the gain cannot be computed when the candidate is chosen, the state contains enough information to predict it in part.}
\end{assumption}

\begin{assumption}[\textcolor{black}{Similar tasks}]\label{ass:transfer}
\textcolor{black}{The relation between the state and the gain $v_i(s)$ is similar on the training tasks and on the task where the selection takes place.}
\end{assumption}

{\color{black}
Assumption~\ref{ass:tasks} makes it possible to measure the gain associated with each candidate, Assumption~\ref{ass:predict} enables learning to predict these gains, and Assumption~\ref{ass:transfer} allows the learned predictions to be transferred to a new task. Assumption~\ref{ass:tasks} is the most demanding, as it requires tasks for which the hidden information is available and utility evaluations can be performed. Although these evaluations are costly, they are carried out once, before any new task is considered, and what is learned from them can then be reused on related tasks. Evaluating the gain of every candidate can itself be too costly, and Section~\ref{sec:salve} describes how it is then approximated.

The setting above encompasses a broad range of selection problems. The candidates can be of any type, provided that each candidate is represented by a vector. Selection can take place in a single round or over multiple rounds. The proposed framework applies to problems in this setting for which Assumptions~\ref{ass:tasks} to~\ref{ass:transfer} hold, and Section~\ref{sec:scope} gives a sequence of questions that decides whether a given problem satisfies them.}
\subsection{The SALVES framework}
\label{sec:salve}
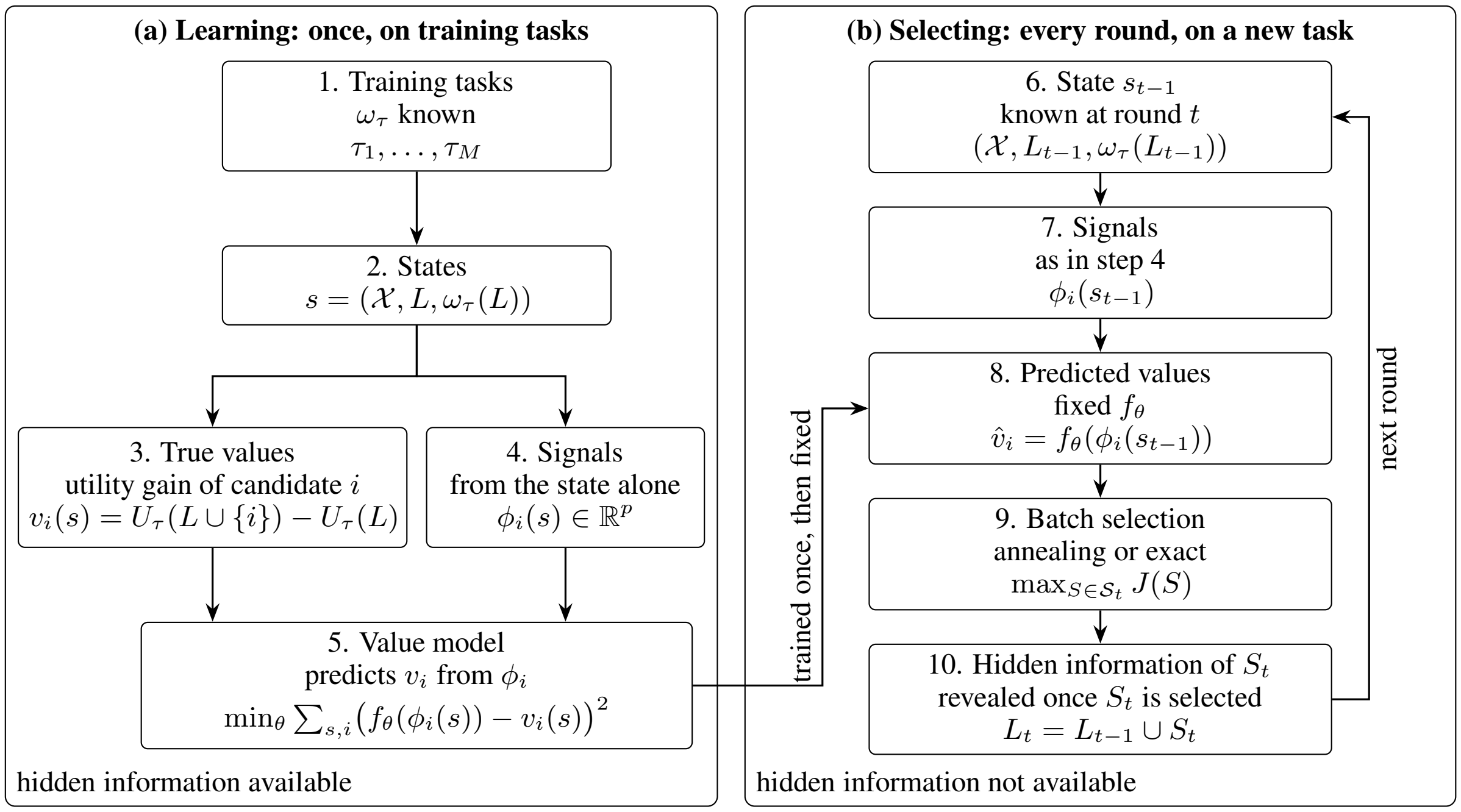
\begin{figure}[!h]
    \centering
    \begin{tikzpicture}[
  font=\footnotesize,
  box/.style={draw, rounded corners=2pt, align=flush center,
              inner sep=3pt, text width=#1},
  box/.default=0.5*\panelw,
  arrow/.style={-{Stealth[length=2mm]}, semithick},
  frame/.style={rounded corners=3pt},
  ptitle/.style={font=\footnotesize\bfseries, anchor=south},
]
\pgfmathsetlengthmacro{\panelw}{0.485*\textwidth}
\pgfmathsetlengthmacro{\panelgap}{3mm}
\pgfmathsetlengthmacro{\halfw}{0.425*\panelw+1mm+6pt}

\begin{scope}[local bounding box=all]
\begin{scope}[local bounding box=panela]
  \node[box] (train_t)
    {1. Training tasks\\ $\omega_\tau$ known\\ $\tau_1,\dots,\tau_M$};
  \node[box, below=8mm of train_t] (state)
    {2. States\\ $s=(\mathcal{X}, L, \omega_\tau(L))$};

  \node[box=0.50*\panelw, minimum height=13mm, anchor=north east]
    (TrueV) at ([shift={(-1mm,-11mm)}]state.south)
    {3. True values\\ utility gain of candidate $i$\\
     $v_i(s)=U_\tau(L\cup\{i\})-U_\tau(L)$};
  \node[box=0.35*\panelw, minimum height=13mm, anchor=north west]
    (signals) at ([shift={(1mm,-11mm)}]state.south)
    {4. Signals\\ from the state alone\\ $\phi_i(s)\in\mathbb{R}^p$};

  \node[box=0.72*\panelw, anchor=north]
    (Value_m) at ([yshift=-8mm]state.south |- TrueV.south)
    {5. Value model\\ predicts $v_i$ from $\phi_i$\\[1pt]
     $\min_{\theta}\sum_{s,i}\bigl(f_\theta(\phi_i(s))-v_i(s)\bigr)^2$};

  \draw[arrow] (train_t) -- (state);
  \draw[arrow] (state.south) -- ++(0,-5.5mm) -| (TrueV.north);
  \draw[arrow] (state.south) -- ++(0,-5.5mm) -| (signals.north);
  \draw[arrow] (TrueV.south)   -- (TrueV.south   |- Value_m.north);
  \draw[arrow] (signals.south) -- (signals.south |- Value_m.north);
\end{scope}

\begin{scope}[local bounding box=panelb,
              node distance=3.6mm, box/.default=0.6*\panelw]
  \node[box, anchor=north] (s_prev)
    at ([xshift=\halfw+8pt+\panelgap]panela.east |- train_t.north)
    {6. State $s_{t-1}$\\ known at round $t$\\
     $(\mathcal{X}, L_{t-1}, \omega_\tau(L_{t-1}))$};
  \node[box, below=of s_prev] (sig_b)
    {7. Signals\\ as in step 4\\ $\phi_i(s_{t-1})$};
  \node[box, below=of sig_b] (pred)
    {8. Predicted values\\ fixed $f_\theta$\\
     $\hat v_i=f_\theta(\phi_i(s_{t-1}))$};
  \node[box, below=of pred] (anneal)
    {9. Batch selection\\ annealing or exact\\ $\max_{S\in\mathcal{S}_t} J(S)$};
  \node[box, below=of anneal] (reveal)
    {10. Hidden information of $S_t$\\ revealed once $S_t$ is selected\\
     $L_t=L_{t-1}\cup S_t$};

  \draw[arrow] (s_prev) -- (sig_b);
  \draw[arrow] (sig_b)  -- (pred);
  \draw[arrow] (pred)   -- (anneal);
  \draw[arrow] (anneal) -- (reveal);
  \draw[arrow] (reveal.east) -- ++(4mm,0) |- (s_prev.east)
    node[pos=0.25, rotate=90, anchor=north, inner sep=2pt] {next round};
  \path ([xshift=-\halfw]s_prev.center) -- ([xshift=\halfw]s_prev.center);
\end{scope}
\end{scope}

\node[ptitle] at ([yshift=0.5mm]panela.north |- all.north)
  {(a) Learning: once, on training tasks};
\node[ptitle] at ([yshift=0.5mm]panelb.north |- all.north)
  {(b) Selecting: every round, on a new task};

\coordinate (aSW) at ([shift={(-4pt,-16pt)}]panela.west |- all.south);
\coordinate (aNE) at ([shift={(4pt,6mm)}]panela.east |- all.north);
\coordinate (bSW) at ([shift={(-4pt,-16pt)}]panelb.west |- all.south);
\coordinate (bNE) at ([shift={(4pt,6mm)}]panelb.east |- all.north);
\coordinate (aNW) at (aSW |- aNE);
\coordinate (bNW) at (bSW |- bNE);
\draw[frame] (aSW) rectangle (aNE);
\draw[frame] (bSW) rectangle (bNE);

\node[anchor=north west] at ([yshift= 5mm] aSW) {hidden information available};
\node[anchor=north west] at ([yshift= 5mm] bSW) {hidden information not available};

\draw[arrow] (Value_m.east) -| ([xshift=-5mm]pred.west) -- (pred.west)
node[pos=-0.15, rotate=90, anchor=south east, inner sep=0pt] {trained once, then fixed};
\end{tikzpicture}
    \caption{Overview of SALVES.}
    \label{fig:overview}
\end{figure}

SALVES is a framework that builds a selection rule from one idea. The gain $v_i(s)$ of Equation~\eqref{eq:gain} cannot be computed on the task where the selection takes place, but it can be computed on training tasks. A model is therefore trained to predict this gain from quantities available during selection, and its predictions are used in place of the unknown gain. The framework has two phases, illustrated in Figure~\ref{fig:overview}. The learning phase is carried out once. The gains are computed on the training tasks and the model is fitted, corresponding to steps 1 to 5 in the figure. The selection phase is then repeated at each round on a new task. The fixed model predicts the gain of each candidate that has not yet been selected. A batch that combines high predicted gains with diversity is then selected, by simulated annealing or exactly when the objective allows it, corresponding to steps 6 to 10. The rest of this subsection describes these steps in general terms, and Section~\ref{sec:applications} specifies them for each application.

The learning phase starts by producing examples on the training tasks, which corresponds to steps 1 to 3 of Figure~\ref{fig:overview}. An example consists of a state, a candidate not yet selected in this state, and the gain of this candidate. On each training task, the selection is run for several rounds with a simple selection rule, for instance a random choice of the batches, and each round provides one state $s=(\mathcal{X},L,\omega_\tau(L))$. When the selection has a single round, the states are the initial states of the training tasks. For every state, the gain $v_i(s)$ is then computed for the candidates not yet selected, or for a random sample of them when they are too many. This is possible because the hidden information of a training task is known. The gain computed in this way is called the true value of the candidate. Computing it exactly requires one evaluation of the utility per candidate, which is costly. When this cost is too high, the exact gain is replaced by a quantity that is cheaper to compute, either an approximation of the gain or a simpler indicator of the usefulness of the candidate.

The true value cannot be computed during the selection, so each candidate is also described by quantities that are available at that stage. These quantities are called signals and correspond to step 4 of Figure~\ref{fig:overview}. For a state $s$ and a candidate $i$ not yet selected, they form a vector $\phi_i(s)\in\mathbb{R}^p$, where $p$ is the number of signals. Two requirements guide their choice.
\begin{enumerate}
  \item A signal must be computable from the state alone, so that it never uses hidden information that has not been revealed.
  \item A signal must have the same meaning on every task, whatever the number of candidates $N$ and the number $d$ of values that describe them, so that a model trained on the training tasks can be applied to a new task.
\end{enumerate}
Typical signals describe what a model fitted on the revealed information predicts for the candidate, how the candidate relates to the candidates already selected, how it relates to the other candidates, and how far the selection has progressed. The signals of each application are listed in Section~\ref{sec:applications}. \textcolor{black}{The second requirement gives a precise meaning to Assumption~\ref{ass:transfer}. A new task is similar to the training tasks when the relation between the signals of a candidate and its gain is the same on the new task as on the training tasks.}

\textcolor{black}{Before the value model is introduced, it is shown that the true value cannot be estimated from a model fitted on the revealed information alone, and must therefore be learned on the training tasks.} Consider the case where the hidden information of a candidate is a class label. The label of candidate $i$ is denoted by $y_i\in\{1,\dots,K\}$, where $K$ is the number of classes. A classifier fitted on the revealed information assigns to candidate $i$ a vector of predicted class probabilities, denoted by $\hat q_i(s)\in\mathbb{R}^K$. The true class probabilities of the candidate, which are unknown, are denoted by $q_i\in\mathbb{R}^K$. Assume that the classifier is trained by minimising a regularised cross-entropy loss. The utility is defined as the negative cross-entropy loss of the trained classifier on a fixed set of labelled examples, called the reference set. The loss is used rather than the accuracy because the result requires a utility that varies smoothly with the parameters of the classifier. \textcolor{black}{The influence function \citep{Ref2} then gives an approximation of the gain of Equation~\eqref{eq:gain} to first order. This approximation is called the first-order gain and is denoted by $\tilde v_i(s)$, so that $v_i(s)\approx\tilde v_i(s)$, with}
\begin{equation}
\tilde v_i(s)\;=\;a_i(s)^\top\big(\hat q_i(s)-e_{y_i}\big),
\label{eq:firstorder}
\end{equation}
where $e_{y}\in\mathbb{R}^K$ is the vector with a one in position $y$ and zeros elsewhere, and $a_i(s)\in\mathbb{R}^K$ is a vector that depends on the state, on the description $x_i$ of the candidate and on the reference set, but not on the label $y_i$.

\begin{proposition}[\textcolor{black}{Expected first-order gain}]\label{prop:expected}
\textcolor{black}{Let the state $s$ be fixed and let the label $y_i$ of candidate $i$ be drawn from its true class probabilities $q_i$. The first-order gain of Equation~\eqref{eq:firstorder} satisfies}
\begin{equation}
\mathbb{E}\big[\tilde v_i(s)\big]\;=\;a_i(s)^\top\big(\hat q_i(s)-q_i\big).
\label{eq:expected}
\end{equation}
\textcolor{black}{Consequently, the expected first-order gain is zero when the predicted probabilities equal the true ones, that is, when $\hat q_i(s)=q_i$. Moreover, replacing the unknown probabilities $q_i$ by the predicted probabilities $\hat q_i(s)$ in Equation~\eqref{eq:expected} gives zero for every candidate.}
\end{proposition}

\noindent\textcolor{black}{\textbf{Proof.} In Equation~\eqref{eq:firstorder}, only the label $y_i$ is random, since $a_i(s)$ and $\hat q_i(s)$ are computed from the state. The label takes the value $k$ with a probability equal to the $k$-th entry of $q_i$, so the expectation of $e_{y_i}$ is $q_i$, which gives Equation~\eqref{eq:expected}. The right-hand side of Equation~\eqref{eq:expected} is zero when $\hat q_i(s)=q_i$, and also when $q_i$ is replaced by $\hat q_i(s)$.} \qed

\textcolor{black}{On average, a candidate is therefore useful only where the classifier is wrong about the probabilities. The expected gain cannot be evaluated during the selection, since the true probabilities are unknown, and the classifier alone therefore cannot tell which candidates are useful.} What can be learned on the training tasks, where the labels are known, is how the gain relates to quantities that are available during the selection. This is the role of the value model.

The value model is the function that replaces the unknown gain during the selection, and it corresponds to step 5 of Figure~\ref{fig:overview}. It is denoted by $f_\theta$, where $\theta$ collects its parameters, and it maps the signals of one candidate to a real number. It is trained on the examples produced on the training tasks, by making its output as close as possible to the true value,
\begin{equation}
\min_{\theta}\;\sum_{s}\sum_{i}\Big(f_\theta\big(\phi_i(s)\big)-v_i(s)\Big)^2,
\label{eq:training}
\end{equation}
where the first sum runs over the states generated on the training tasks and the second over the candidates for which the true value was computed. The gains become smaller as the selection progresses, so the true values are standardised within each state before the fit, by subtracting their mean and dividing by their standard deviation over the candidates of that state. Without this step, the first states would dominate the sum. When the true value is a binary indicator rather than a gain, the squared error is replaced by the cross-entropy, and the output of the value model is the predicted probability that the candidate is useful. Any regression or classification model can be used, and the model retained for each application is given in Section~\ref{sec:applications}. However, two choices are common to all applications.
\begin{enumerate}
  \item The value model receives the signals of one candidate at a time, so that the cost of applying it grows in proportion to the number of candidates.
  \item Some training tasks are kept out of the fit and used to stop the training and compare models. Entire tasks, rather than individual candidates, are kept out because the candidates within a task share the same hidden information. The comparison uses a measure that reflects how the value model is used. For a gain, this measure is the rank correlation between the model outputs and the true values, computed within each state and averaged over states. For a binary indicator, it is the area under the receiver operating characteristic curve.
\end{enumerate}
Once the training is finished, the parameters $\theta$ are fixed. On a new task, the output $\hat v_i=f_\theta(\phi_i(s_{t-1}))$ is called the predicted value of candidate $i$ at round $t$.

The selection step turns the predicted values into a batch. At round $t$, the signals and the predicted values are computed for every candidate not yet selected, which corresponds to steps 6 to 8 of Figure~\ref{fig:overview}. Selecting the $B_t$ candidates with the highest predicted values is not sufficient. The true value of Equation~\eqref{eq:gain} is the gain of adding one candidate alone, and the gains of several candidates do not add up when these candidates carry the same information. Candidates with similar descriptions have similar predicted values, so the highest ones are often near-duplicates of each other. The batch is therefore required to be diverse as well. The quality of a batch $S$ is measured by an objective $J$ that adds the predicted values of its candidates and a diversity term,
\begin{equation}
J(S)\;=\;\sum_{i\in S}\hat v_i\;+\;\lambda\,D(S),
\label{eq:objective}
\end{equation}
where $D(S)$ measures the diversity of the batch and $\lambda\ge 0$ is the weight given to it. The batch selected at round $t$ is an admissible batch with the highest objective,
\begin{equation}
S_t\in\argmax_{S\in\mathcal{S}_t}\;J(S),
\label{eq:selection}
\end{equation}
where $\mathcal{S}_t$ is the set of admissible batches, that is, of all sets of $B_t$ candidates not yet selected.

Diversity can be imposed in two ways.
\begin{enumerate}
  \item With a diversity term, $D(S)$ measures how far the candidates of the batch are from each other and from the candidates already selected. The weight $\lambda$ sets the balance between predicted value and diversity, and every batch of $\mathcal{S}_t$ remains admissible.
  \item With a group constraint, similar candidates are first gathered into $G\ge B_t$ groups, and $\mathcal{S}_t$ is restricted to the batches that contain at most one candidate per group. Diversity is then ensured by the constraint, and $\lambda=0$.
\end{enumerate}
The diversity term used in this paper is the mean distance from each candidate of the batch to its nearest neighbour among the other candidates of the batch and the candidates already selected,
\begin{equation}
D(S)=\frac{1}{B_t}\sum_{i\in S}\;\min_{j\in(S\setminus\{i\})\cup L_{t-1}}\delta(i,j),
\label{eq:spread}
\end{equation}
where $\delta(i,j)$ is a dissimilarity between the descriptions of candidates $i$ and $j$, given for each application in Section~\ref{sec:applications}. A candidate close to another candidate of the batch, or to one already selected, adds little to $D(S)$.

With the diversity term, no shortcut exists in general, because $D(S)$ depends on the relationships among the candidates within the batch. The best batch has to be found by a search, which corresponds to step 9 of Figure~\ref{fig:overview}. Simulated annealing \citep{Ref1} is used for this purpose because it requires only that the objective of a batch can be evaluated. The search starts from the $B_t$ candidates with the highest predicted values and modifies the batch one candidate at a time. At each iteration, one candidate of the current batch is replaced by a candidate outside it. The candidate brought in is drawn with a probability that increases with its predicted value, so that the search spends more iterations on promising candidates. The resulting change in the objective $J$ is denoted by $\Delta$. The replacement is accepted with probability
\begin{equation}
\min\big\{1,\;\exp(\Delta/\gamma)\big\},
\label{eq:acceptance}
\end{equation}
where $\gamma>0$ is a parameter called the temperature. An improvement, for which $\Delta$ is positive, is therefore always accepted. The initial temperature is the median of $|\Delta|$ over a sample of random replacements, so that it matches the scale of the objective. It is then multiplied by the same factor at every iteration, so that it reaches one hundredth of its initial value at the last one, and the search ends by accepting improvements only. The batch with the highest objective met during the search is returned. Each iteration only requires the change $\Delta$, which is computed from the predicted values and the diversity of the batch.

Under the group constraint, the optimal batch can be computed exactly without search.

\begin{proposition}[\textcolor{black}{Exact solution under the group constraint}]\label{prop:group}
\textcolor{black}{Under the group constraint, the best candidate of group $g$ is the candidate of this group that has not yet been selected and has the highest predicted value, and its predicted value is denoted by $m_g$. The batch formed by the best candidates of the $B_t$ groups with the highest values of $m_g$ is a solution of Equation~\eqref{eq:selection}. It is obtained with one pass over the candidates and one sorting of the groups, that is, in $O(N+G\log G)$ operations.}
\end{proposition}

\noindent\textcolor{black}{\textbf{Proof.} Under the group constraint, $\lambda=0$, so the objective of a batch is the sum of the predicted values of its candidates. Any admissible batch contains candidates from $B_t$ different groups, and the candidate taken from group $g$ has a predicted value of at most $m_g$. The objective of the batch is therefore at most the sum of the $B_t$ highest values of $m_g$, which is the objective of the batch above.} \qed

The same argument applies when the predicted value is replaced by another candidate-level score, such as the rank of the predicted value combined with one of the signals of the candidate.

The two ways suit different batch sizes. \textcolor{black}{The diversity term is flexible, since two similar candidates can still be selected together if both have high predicted values, but its weight $\lambda$ must be chosen and its search becomes expensive when the batch and the number of candidates are large. The group constraint is stricter, since it never selects two candidates of the same group, but it has no weight to choose and its solution is exact.} The diversity term is therefore used for small batches, and the group constraint for large ones. When neither is used, the batch consists of the $B_t$ candidates with the highest predicted values. The form used in each application is given in Section~\ref{sec:applications}.

Step 10 of Figure~\ref{fig:overview} reveals the hidden information of the selected batch and completes the round. The selected set becomes $L_t=L_{t-1}\cup S_t$, and the new state is $s_t=(\mathcal{X},L_t,\omega_\tau(L_t))$. The next round starts from this state, so that the newly revealed information guides the following selection. Algorithm~\ref{alg:round} summarises one round with the diversity term. Under the group constraint or without diversity, lines 3 to 10 are replaced by the exact batch.

\begin{algorithm}[!h]
\caption{One round of SALVES on a new task}
\label{alg:round}
\begin{algorithmic}[1]
\Require state $s_{t-1}$, value model $f_\theta$, batch size $B_t$
\State compute the signals $\phi_i(s_{t-1})$ of every candidate $i$ not yet selected
\State compute the predicted values $\hat v_i=f_\theta(\phi_i(s_{t-1}))$
\State $S\gets$ the $B_t$ candidates with the highest predicted values
\Repeat
     \State $S'\gets S$ with one candidate replaced, preferring a high $\hat v_i$
  \State $\Delta\gets J(S')-J(S)$
  \State with probability $\min\{1,\exp(\Delta/\gamma)\}$, set $S\gets S'$
  \State decrease the temperature $\gamma$
\Until{the number of iterations is reached}
\State $S_t\gets$ the batch with the highest objective met during the search
\State reveal the hidden information of $S_t$ and set $L_t\gets L_{t-1}\cup S_t$
\Ensure state $s_t=(\mathcal{X},L_t,\omega_\tau(L_t))$
\end{algorithmic}
\end{algorithm}

The computational cost of the framework is divided between its learning and selection phases. The learning phase carries the main cost because it requires computing the true values. Each true value requires the hidden information and, when computed exactly, one evaluation of the utility. Thus, for $E$ examples, computing the true values requires $O(E)$ utility evaluations. This cost is incurred once, before the framework is applied to new tasks.

The selection phase is comparatively light. At each round, the signals are computed and the value model is applied once to each candidate that has not yet been selected, giving a cost of order $O(N)$. The search then operates on these precomputed quantities, and its cost grows with its number of iterations. Under the group constraint, the exact solution of Proposition~\ref{prop:group} costs only $O(N+G\log G)$, and without diversity the selection reduces to a sort. The utility is never evaluated during batch selection. The only costly operation within a round is therefore the one imposed by the problem itself, namely the revelation of the hidden information associated with the selected batch.

Classical selection rules are special cases of the framework, obtained when the value is fixed by hand instead of learned. When the predicted value is replaced by a single signal and no diversity is used, the framework selects the candidates that are best according to this signal, such as the smallest margin of the classifier in uncertainty sampling \citep{Ref3} or the largest correlation with the response in filter methods \citep{Ref18}. When the value is replaced by the typicality of a candidate under the group constraint, it selects one typical candidate per group, close to TypiClust \citep{Ref9}. When the value is constant and only the diversity term is kept, it selects candidates far from each other and from those already selected, close to the core-set rule \citep{Ref8}. What SALVES adds to these rules is a value learned on training tasks, which combines several signals and adapts their weights to the state.

The selection step can be used in two ways, which differ in what a candidate is. In the direct use, described so far, the candidates are the items of the final solution, such as images or bands, and the selected set is the solution itself. In the move-filter use, the solution is built by a local search, such as simulated annealing, which repeatedly proposes a small modification of the current solution, evaluates the criterion of the modified solution, and keeps or rejects the modification. When each evaluation is costly, the number of evaluations is the budget, and SALVES decides which modifications are worth evaluating. At each step of the search, a few modifications are proposed at random, the value model scores each of them, and only the one with the highest predicted value is evaluated. The search then keeps or rejects it by its usual rule. In the terms of Section~\ref{sec:problem}, the candidates are the proposed modifications, the batch contains a single modification, and the hidden information of a modification is its effect on the criterion, which is known only once it has been evaluated. The proposals remain random, so that the search keeps the randomness it needs to escape poor solutions. The value model is trained as in the direct use, on modifications proposed and evaluated during searches on training tasks, and its signals satisfy the same two requirements.
\subsection{Applicability of SALVES}
\label{sec:scope}
This subsection explains why a learned selection rule is needed, defines the class of problems to which SALVES applies, gives a sequence of questions that decides whether and in which form the framework can be used for a given problem, and identifies the settings in which SALVES is expected to outperform classical selection rules.

\textcolor{black}{As discussed in Section~\ref{sec:related}, the selection rules used in practice rely on one fixed criterion, such as the uncertainty of a classifier or the correlation between a variable and the response, and no single criterion is the strongest across tasks \citep{Ref18, ref19b, ref20}. Systematic comparisons also reveal a large margin for improvement.} In one of these comparisons, the rules are measured against an oracle, a rule that cannot be used in practice because it looks at the labels of the candidates before choosing them. Among a random sample of candidates, the oracle adds the one that improves the classifier most, and it outperforms all the existing rules by at least 5\% \citep{ref20}. This gap shows how much is lost by choosing without knowing which candidates truly improve the classifier. A selection rule learned on tasks where the hidden information is known addresses both observations. It is not tied to a single criterion, and it is trained to predict, without the labels, the gain that the oracle computes with them.

\textcolor{black}{The problem of Section~\ref{sec:problem} defines a broad class of problems.}

\begin{definition}[\textcolor{black}{Problem class}]\label{def:class}
\textcolor{black}{Budgeted selection under hidden information over a distribution of tasks is the class of problems that satisfy three conditions. First, a subset of $n<N$ candidates is selected, in one or several rounds. Second, the utility of the subset depends on hidden information, which is revealed only for the candidates that are selected. Third, this information is known on other tasks of the same family, which have already been solved.}
\end{definition}

\textcolor{black}{The hidden information may be obtained by labelling, measuring, testing or evaluating the candidates.} The usefulness of a candidate therefore has to be anticipated on a new task. This distinguishes the class from classical combinatorial optimisation, where the objective of a candidate solution can be computed whenever the search needs it. The class covers problems from several communities: labelling data for a classifier in machine learning, choosing spectral bands or wavelengths in remote sensing and chemometrics, choosing variables or biomarkers in statistics and bioinformatics, choosing molecules to test in drug discovery, placing sensors in environmental monitoring, and choosing which modifications of a design to evaluate in the design of experiments. Four of these problems are studied in this paper, namely active learning, band selection, input selection and the construction of designs.

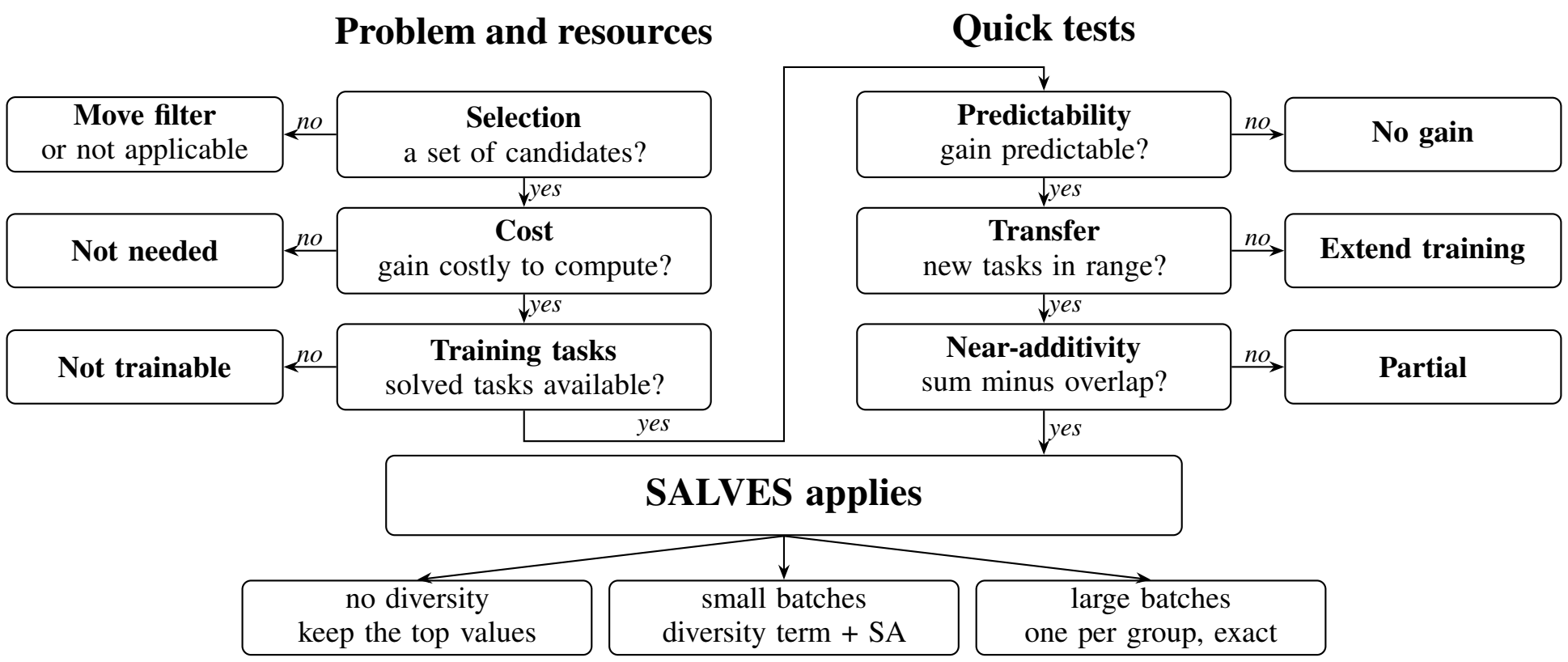
\begin{figure}[!h]
    \centering
\begin{tikzpicture}[
    quest/.style={draw=black, fill=white, line width=0.4pt,
                  rounded corners=2pt, text width=\qw cm,
                  minimum height=\qh cm, inner sep=1.5pt,
                  align=center, font=\scriptsize},
    verdict/.style={draw=black, fill=white, line width=0.4pt,
                    rounded corners=2pt, text width=\vw cm,
                    minimum height=\vh cm, inner sep=1.5pt,
                    align=center, font=\scriptsize},
    form/.style={verdict, text width=\fw cm},
    colhead/.style={inner sep=0pt, font=\footnotesize\bfseries},
    yesno/.style={inner sep=1pt, font=\tiny\itshape},
    flow/.style={-{Stealth[length=3.5pt]}, line width=0.4pt},
  ]

  \def\qw{2.95}
  \def\qh{0.7}
  \def\vw{2.15}
  \def\vh{0.6}
  \def\fw{2.75}
  \def\fgap{3.0}
  \def\okw{6.4}
  \def\pitch{0.95}
  \def\colr{4.25}
  \def\side{3.1}
  \def\ylink{0.55}
  \def\yok{-2.95}
  \def\yform{-3.95}

  \node[colhead] at (0,0.85) {Problem and resources};
  \node[colhead] at (\colr,0.85) {Quick tests};

  \node[quest] (q0) at (0,0) {{\bfseries Selection}\\a set of candidates?};
  \node[quest] (q1) at (0,-\pitch) {{\bfseries Cost}\\gain costly to compute?};
  \node[quest] (q2) at (0,{-2*\pitch}) {{\bfseries Training tasks}\\solved tasks available?};

  \node[quest] (q3) at (\colr,0) {{\bfseries Predictability}\\gain predictable?};
  \node[quest] (q4) at (\colr,-\pitch) {{\bfseries Transfer}\\new tasks in range?};
  \node[quest] (q5) at (\colr,{-2*\pitch}) {{\bfseries Near-additivity}\\sum minus overlap?};

  \foreach \qa/\qb in {0/1, 1/2, 3/4, 4/5} {
    \draw[flow] (q\qa) -- node[yesno, right] {yes} (q\qb);
  }
  \draw[flow] (q2.south) -- ++(0,-0.25) -| node[yesno, pos=0.25, above] {yes}
    ({\colr/2},\ylink) -| (q3.north);

  \node[verdict] (v0) at (-\side,0) {{\bfseries Move filter}\\or not applicable};
  \node[verdict] (v1) at (-\side,-\pitch) {{\bfseries Not needed}};
  \node[verdict] (v2) at (-\side,{-2*\pitch}) {{\bfseries Not trainable}};
  \node[verdict] (v3) at ({\colr+\side},0) {{\bfseries No gain}};
  \node[verdict] (v4) at ({\colr+\side},-\pitch) {{\bfseries Extend training}};
  \node[verdict] (v5) at ({\colr+\side},{-2*\pitch}) {{\bfseries Partial}};
  \foreach \qk in {0,...,5} {
    \draw[flow] (q\qk) -- node[yesno, above] {no} (v\qk);
  }

  \node[verdict, text width=\okw cm, minimum height=0.65cm, font=\footnotesize]
    (ok) at ({\colr/2},\yok) {{\bfseries SALVES applies}};
  \draw[flow] (q5) -- node[yesno, right] {yes} (q5 |- ok.north);
  \node[form] (m0) at ({\colr/2-\fgap},\yform) {no diversity\\keep the top values};
  \node[form] (m1) at ({\colr/2},\yform) {small batches\\diversity term + SA};
  \node[form] (m2) at ({\colr/2+\fgap},\yform) {large batches\\one per group, exact};
  \foreach \qk in {0,1,2} {
    \draw[flow] (ok.south) -- (m\qk.north);
  }

\end{tikzpicture}
    \caption{Applicability of SALVES.}
    \label{fig:applicability}
\end{figure}

Whether a new problem falls within the scope of SALVES, and which form of the framework applies to it, can be determined through a short sequence of questions, summarised in Figure~\ref{fig:applicability}. Two of these questions can be answered from the problem statement alone, and the third from the resources available for the problem.

\begin{enumerate}
    \item Does a solution consist of a set of candidates, such that its utility depends only on which candidates are selected, regardless of their order, assigned positions or group memberships? This question does not exclude a problem but determines the form in which SALVES is used. If the answer is yes, SALVES selects the candidates directly. Routing and scheduling problems, in which the contribution of a candidate depends on its neighbours in a sequence, do not satisfy this condition, nor do partitioning and assignment problems. Such problems remain within reach when their solution is improved through local modifications whose evaluation is costly. SALVES then selects which modifications to evaluate, through the move-filter use described in Section~\ref{sec:salve}. Otherwise, they fall outside its scope.
    \item Is the gain associated with a candidate expensive to compute, or difficult to approximate accurately, for a new task? If gains can be computed or approximated accurately at low cost, SALVES is unnecessary, as they can be obtained directly rather than predicted. The travelling salesman problem illustrates both situations in which the answer is negative. The effect of exchanging two edges in a tour can be computed as cheaply as any signal, while the suitability of an edge for inclusion in a good tour can be approximated accurately using a relaxation of the problem \citep{ref21}. In the applications of this paper, by contrast, the gain cannot be obtained without the hidden information, since it requires the labels of the candidates, many labelled samples or a costly evaluation of the design.
    \item Are solved tasks from the same problem family available, with the hidden information fully known, so that the gain defined in Equation~\eqref{eq:gain} can be computed for many states and candidates at an acceptable cost? These tasks can be obtained from labelled data, simulators, synthetic generators or past campaigns. Their availability is Assumption~\ref{ass:tasks}. If the cost of computing the gains is prohibitive, the gain can be approximated or replaced by a simpler indicator of usefulness, as discussed in Section~\ref{sec:salve}. Without suitable training tasks, however, there is no basis for learning the value model, and SALVES cannot be trained.
\end{enumerate}

The final three questions concern the relationship between the signals and the gain. They can be answered through preliminary tests on the training tasks, before applying SALVES to any new task.

\begin{enumerate}
    \setcounter{enumi}{3}
    \item Can the gain be predicted, at least partially, from signals satisfying the two requirements specified in Section~\ref{sec:salve}? This question corresponds to Assumption~\ref{ass:predict}. \textcolor{black}{It is assessed on held-out training tasks, with the measures described in Section~\ref{sec:salve}.} Performance at chance level indicates that no improvement over classical selection rules can be expected from the value predictions.
    \item Do the new tasks fall within the range covered by the training tasks? This question corresponds to Assumption~\ref{ass:transfer}. It is assessed by comparing the signal distributions across both sets of tasks and by holding out entire groups of training tasks that differ from the others in ways representative of the expected differences in new tasks, such as different classes or sensors. If they do not, the training tasks must be extended to cover them.
    \item Can the gain of a batch be approximated by the sum of its individual candidate gains, adjusted by a penalty that increases with candidate similarity? This approximation is particularly unlikely to hold when candidates are useful only in combination with others, as in problems where variables provide information about the response only jointly. In such cases, assigning a value to each candidate individually cannot capture their joint contribution. Whether the approximation holds can be assessed by comparing the true batch gains with it on the training tasks or by checking whether a model that predicts the gains of entire batches outperforms the combination of individual value predictions and diversity. This assessment also requires a measure of similarity between candidate descriptions, consistent with the measure used in the diversity term. If the approximation fails, SALVES captures only part of the usefulness of the candidates.
\end{enumerate}

When all the answers are positive, SALVES applies directly, and the form of the selection step follows from Section~\ref{sec:salve}. When the value of a candidate already accounts for redundancy, or when the batch contains a single candidate, the candidates with the highest predicted values are selected. Otherwise, the diversity term optimised by simulated annealing suits small batches, and the group constraint, which allows at most one candidate per group, suits large batches.

Applying these questions to the problems discussed above yields three groups. The first comprises problems for which the questions concerning the problem structure and available resources are answered positively and the preliminary tests are expected to succeed. It contains the problems listed above other than the construction of designs, together with the choice of examples to annotate for a language model and the selection of training data. In these problems, large numbers of training tasks can be generated, and the candidates interact mainly through redundancy. The second group comprises problems to which SALVES applies in a restricted form. When candidates are useful mainly in combination, as with variables that contribute information only jointly, individual value predictions capture only part of their usefulness. When a solution is not a set of candidates but is improved by local modifications whose evaluation is costly, as in the construction of designs, SALVES is used as a move filter within the search. The third group comprises problems to which SALVES does not apply or for which it offers no expected improvement. Routing, scheduling, partitioning and assignment problems do not satisfy the selection formulation, while problems in which candidate usefulness can be computed or accurately approximated at low cost do not require a learned value model. The travelling salesman problem belongs to this third group, even when reformulated as selecting the connections available to a solver. Although a value model trained on inexpensive signals may transfer to larger and different instances, it is not expected to improve on the candidate lists provided by classical relaxations.

Among the problems to which SALVES applies, its advantage over the classical rules is expected to be largest under four conditions, each of which follows from the construction of the framework:
\begin{itemize}
    \item The information available on the new task is scarce, with few labels, few labelled samples or many more variables than samples. The classical rules then rely on a statistic estimated on the new task alone, which is noisy, whereas the value model brings what was learned on many training tasks.
    \item No single criterion is reliable across tasks, because the most useful candidates change with the budget or with the structure of the task. A fixed criterion is then right in some regimes and wrong in others, whereas the value model combines several criteria and adapts their weights to the state, in particular to the progress of the selection.
    \item The usefulness of a candidate cannot be inferred from the state by a classical computation, even approximately. This is the case when the expected gain depends on errors that the model fitted on the revealed information cannot see, as shown by Proposition~\ref{prop:expected}.
    \item The batches are large or the candidates redundant, so that the diversity of a batch matters as much as the values of its candidates.
\end{itemize}
Conversely, when a classical rule is already close to the best choice, as uncertainty sampling once many labels are available, or when the usefulness of a candidate can be estimated cheaply, SALVES can match the best rule but cannot be expected to surpass it.
\subsection{Applications}
\label{sec:applications}
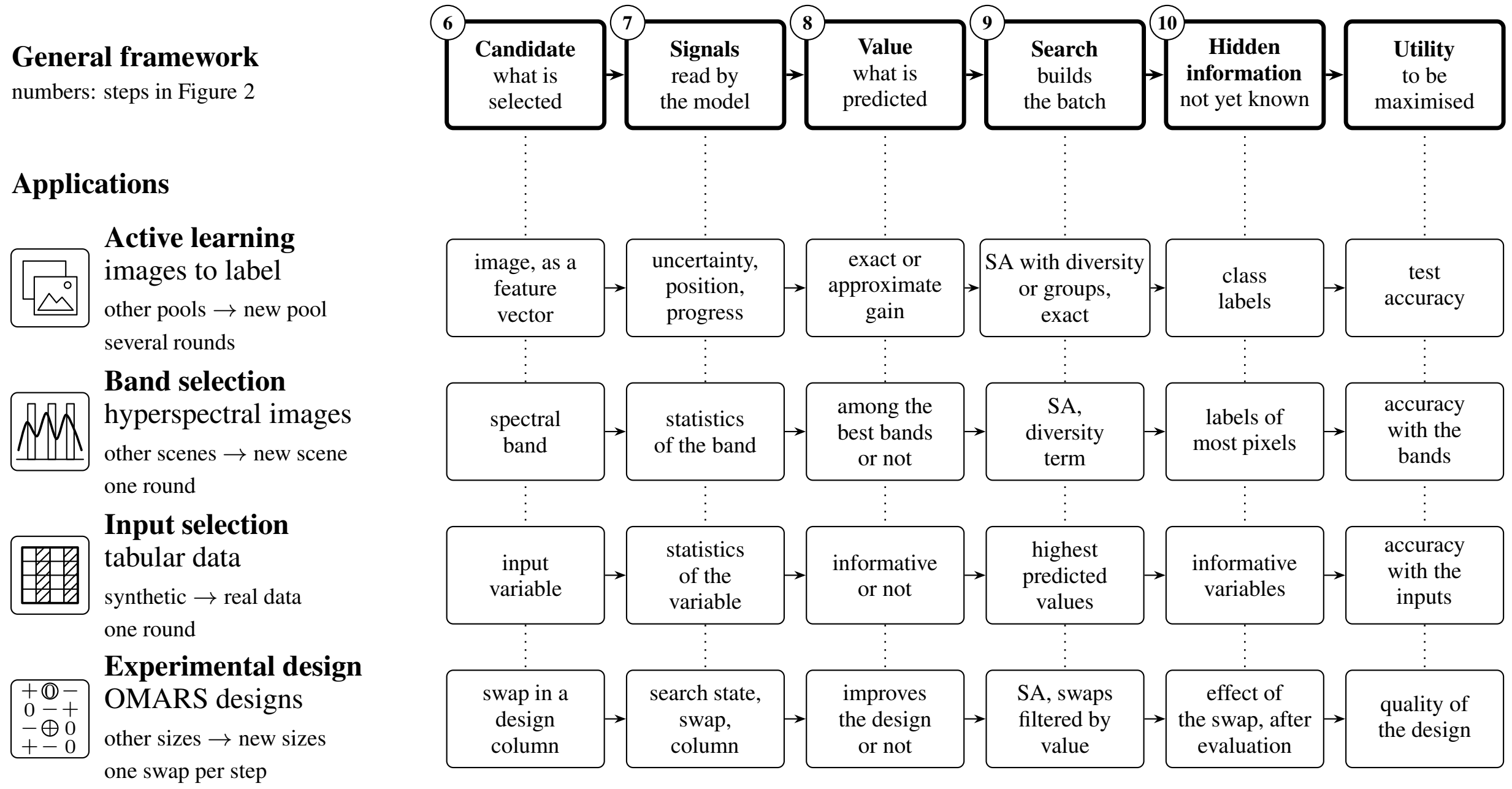
\begin{figure}[!h]
    \centering
\begin{tikzpicture}[
    cell/.style={draw=black, fill=white, line width=0.4pt,
                 rounded corners=2pt, minimum width=\colw cm,
                 minimum height=\cellh cm, inner sep=1.5pt,
                 align=center, font=\scriptsize},
    head/.style={cell, line width=1.3pt, minimum height=\headh cm},
    stepno/.style={circle, draw=black, fill=white, line width=0.4pt,
                   minimum size=3.7mm, inner sep=0pt, font=\tiny\bfseries},
    rowlabel/.style={anchor=west, align=left, inner sep=0pt,
                     font=\footnotesize},
    guide/.style={dotted, line width=0.5pt},
    headarrow/.style={-{Stealth[length=4.5pt]}, line width=0.8pt},
    rowarrow/.style={-{Stealth[length=3.5pt]}, line width=0.4pt},
    icon/.style={line width=0.4pt, line join=round, line cap=round},
  ]

  \def\colw{1.7}
  \def\colgap{0.24}
  \def\xzero{4.7}
  \def\cellh{1.05}
  \def\headh{1.15}
  \def\rowone{-2.3}
  \def\rowpitch{1.55}
  \def\colx#1{\xzero+\colw/2+#1*(\colw+\colgap)}
  \def\rowy#1{\rowone-#1*\rowpitch}

  \def\approw#1#2#3#4#5#6#7{%
    \node[cell] (r#1-0) at ({\colx{0}},{\rowy{#1}}) {#2};
    \node[cell] (r#1-1) at ({\colx{1}},{\rowy{#1}}) {#3};
    \node[cell] (r#1-2) at ({\colx{2}},{\rowy{#1}}) {#4};
    \node[cell] (r#1-3) at ({\colx{3}},{\rowy{#1}}) {#5};
    \node[cell] (r#1-4) at ({\colx{4}},{\rowy{#1}}) {#6};
    \node[cell] (r#1-5) at ({\colx{5}},{\rowy{#1}}) {#7};
    \foreach \ca [evaluate=\ca as \cb using int(\ca+1)] in {0,...,4} {
      \draw[rowarrow] (r#1-\ca) -- (r#1-\cb);
    }
  }

  \def\applabel#1#2#3#4#5{%
    \node[rowlabel] at (1.0,{\rowy{#1}})
      {{\small\bfseries #2}\\#3\\[1.5pt]{\scriptsize #4}\\{\scriptsize #5}};
    \draw[icon, rounded corners=2pt]
      (0,{\rowy{#1}-0.42}) rectangle (0.84,{\rowy{#1}+0.42});
  }

  \foreach \ca in {0,...,5} {
    \draw[guide] ({\colx{\ca}},{-\headh/2}) -- ({\colx{\ca}},{\rowy{3}+\cellh/2});
  }

  \node[anchor=west, align=left, inner sep=0pt, font=\footnotesize] at (0,0)
    {{\small\bfseries General framework}\\
     {\scriptsize numbers: steps in Figure~\ref{fig:overview}}};

  \node[head] (h0) at ({\colx{0}},0)
    {{\bfseries Candidate}\\{\scriptsize what is}\\{\scriptsize selected}};
  \node[head] (h1) at ({\colx{1}},0)
    {{\bfseries Signals}\\{\scriptsize read by}\\{\scriptsize the model}};
  \node[head] (h2) at ({\colx{2}},0)
    {{\bfseries Value}\\{\scriptsize what is}\\{\scriptsize predicted}};
  \node[head] (h3) at ({\colx{3}},0)
    {{\bfseries Search}\\{\scriptsize builds}\\{\scriptsize the batch}};
  \node[head] (h4) at ({\colx{4}},0)
    {{\bfseries Hidden}\\{\bfseries information}\\{\scriptsize not yet known}};
  \node[head] (h5) at ({\colx{5}},0)
    {{\bfseries Utility}\\{\scriptsize to be}\\{\scriptsize maximised}};
  \foreach \ca [evaluate=\ca as \cb using int(\ca+1)] in {0,...,4} {
    \draw[headarrow] (h\ca) -- (h\cb);
  }
  \foreach \ca/\stepnum in {0/6, 1/7, 2/8, 3/9, 4/10} {
    \node[stepno] at ([xshift=1pt,yshift=-1pt]h\ca.north west) {\stepnum};
  }

  \node[anchor=west, inner sep=0pt, font=\small\bfseries] at (0,-1.2)
    {Applications};

  \applabel{0}{Active learning}{images to label}
           {other pools $\to$ new pool}{several rounds}
  \approw{0}{image, as a\\feature\\vector}
            {uncertainty,\\position,\\progress}
            {exact or\\approximate\\gain}
            {SA with diversity\\or groups,\\exact}
            {class\\labels}
            {test\\accuracy}

  \applabel{1}{Band selection}{hyperspectral images}
           {other scenes $\to$ new scene}{one round}
  \approw{1}{spectral\\band}
            {statistics\\of the band}
            {among the\\best bands\\or not}
            {SA,\\diversity\\term}
            {labels of\\most pixels}
            {accuracy\\with the\\bands}

  \applabel{2}{Input selection}{tabular data}
           {synthetic $\to$ real data}{one round}
  \approw{2}{input\\variable}
            {statistics\\of the\\variable}
            {informative\\or not}
            {highest\\predicted\\values}
            {informative\\variables}
            {accuracy\\with the\\inputs}

  \applabel{3}{Experimental design}{OMARS designs}
           {other sizes $\to$ new sizes}{one swap per step}
  \approw{3}{swap in a\\design\\column}
            {search state,\\swap,\\column}
            {improves\\the design\\or not}
            {SA, swaps\\filtered by\\value}
            {effect of\\the swap, after\\evaluation}
            {quality of\\the design}

  \begin{scope}[shift={(0.42,{\rowy{0}})}, icon]
    \draw (-0.29,-0.12) rectangle (0.17,0.29);
    \draw[fill=white] (-0.17,-0.29) rectangle (0.29,0.12);
    \draw (-0.13,-0.24) -- (-0.01,-0.06) -- (0.07,-0.16) -- (0.13,-0.10)
          -- (0.25,-0.24) -- cycle;
    \draw (0.19,0.03) circle (0.035);
  \end{scope}

  \begin{scope}[shift={(0.42,{\rowy{1}})}, icon]
    \foreach \bx in {-0.2,0.02,0.22} {
      \draw (\bx-0.04,-0.3) rectangle (\bx+0.04,0.3);
    }
    \draw[line width=0.7pt] plot[smooth] coordinates
      {(-0.34,-0.2) (-0.25,0.1) (-0.14,-0.05) (-0.04,0.2) (0.07,-0.08)
       (0.17,0.18) (0.27,0.0) (0.34,-0.2)};
    \draw (-0.36,-0.3) -- (0.36,-0.3);
  \end{scope}

  \begin{scope}[shift={(0.42,{\rowy{2}})}, icon]
    \draw[pattern=north east lines] (-0.15,-0.3) rectangle (0,0.3);
    \draw[pattern=north east lines] (0.15,-0.3) rectangle (0.3,0.3);
    \draw[step=0.15] (-0.3,-0.3) grid (0.3,0.3);
    \draw[line width=0.7pt] (-0.3,-0.3) rectangle (0.3,0.3);
  \end{scope}

  \begin{scope}[shift={(0.42,{\rowy{3}})}, icon]
    \foreach \sym [count=\idx from 0] in
        {+,0,-, 0,-,+, -,+,0, +,-,0} {
      \pgfmathsetmacro{\cx}{-0.22+0.22*mod(\idx,3)}
      \pgfmathsetmacro{\cy}{0.3-0.2*floor(\idx/3)}
      \node[inner sep=0pt, font=\tiny] at (\cx,\cy) {$\sym$};
    }
    \draw (0,0.3) circle (0.085);
    \draw (0,-0.1) circle (0.085);
  \end{scope}

\end{tikzpicture}
    \caption{Task-specific elements for each application.}
    \label{fig:applications}
\end{figure}

SALVES is applied to four selection problems: batch active learning, band selection in hyperspectral images, input selection and the construction of experimental designs. The framework of Section~\ref{sec:salve} is the same in all of them, and only six elements are specific to each problem: the candidate, the signals, the value and the model that predicts it, the search that builds the batch, the hidden information, and the utility. Figure~\ref{fig:applications} summarises these six elements for the four applications, and the paragraphs below describe each application in turn. The datasets on which the applications are evaluated are presented with the results.

In batch active learning, a classifier is trained on a pool of images whose labels are costly to obtain, and the labels are requested in batches, over several rounds, until the budget is spent. The selection starts from a small labelled set with the same number of images in each class, drawn at random. A candidate is an unlabelled image of the pool, described by a vector of features extracted by a pretrained network. It is characterised by $p=7$ signals belonging to three categories: uncertainty, position and progress. Three signals measure the uncertainty of the current classifier, since, by Proposition~\ref{prop:expected}, the expected gain is non-zero only where the predicted probabilities are wrong, and uncertain predictions are wrong more often. The entropy of the predicted class probabilities is high when the probability is spread over many classes. The margin, defined as the difference between the two largest probabilities, is small when the classifier hesitates between two classes, typically near a decision boundary. The largest predicted probability measures the confidence of the classifier in its predicted class. These measures capture different aspects of uncertainty, particularly when the probability is distributed unevenly, and the value model learns how to combine them.

The four other signals describe the position of an image in the pool and the progress of the selection. The distance to the nearest selected image is large when the image lies in a region that is not yet covered by labelled data. The density, defined as the mean distance to the nearest images in the pool, distinguishes typical images, whose labels can inform many similar images, from isolated ones, whose labels inform fewer images. The fraction of the pool already selected and the fraction of rounds completed describe the progress of the selection. They allow the value model to adapt its preferences to the remaining budget, since typical images tend to be more useful when few labels are available, whereas uncertain images become more useful later. The entropy, margin, distance and density are converted to ranks within the state, so that their meaning does not depend on the number of classes or on the feature representation. The three remaining signals keep their original values, because they already have the same meaning across tasks. These seven signals were retained on validation tasks from a set of 38 candidate signals, and their contribution is examined through the ablations in Section~\ref{sec:ablations}.

The value is the gain obtained by labelling the image, as in Equation~\eqref{eq:gain}. When the pool is small and the classifier is fast to train, it is computed exactly, as the change in accuracy of the retrained classifier on labelled images kept for this purpose. Otherwise, retraining once per candidate is too costly, and the gain is approximated by the first-order gain of Equation~\eqref{eq:firstorder}, the decrease in loss on a reference set of labelled images. The value model is a multilayer perceptron with two hidden layers of 64 units.

For small batches, diversity is encouraged by the diversity term of Equation~\eqref{eq:spread}, with the cosine distance between the features of two images as the dissimilarity $\delta$, and the batch is found by simulated annealing. For large batches, diversity is imposed by a group constraint. The groups are obtained by $k$-means clustering of the whole pool, with 1.5 groups per image of the batch. Within the groups, the images are compared by a score equal to the rank of the predicted value plus 0.5 times the typicality of the image, the inverse of its mean distance to its nearest neighbours in the same group, also converted to a rank. Typical images represent their group well, and their labels inform the many similar images around them. The batch is then computed exactly, as in Proposition~\ref{prop:group}, without search. When the budget is large, the selection is handed over to margin sampling once enough labels have been selected, since the classifier is then accurate enough for its uncertainty to identify useful images directly. The hidden information is the class label of each image, revealed at the end of the round in which the image is selected. The utility is the accuracy, on a test set, of a logistic regression trained on the features of the selected images.

In band selection, a hyperspectral image records each pixel at many wavelengths, called bands, and a few bands must be chosen so that a classifier trained on a small sample of labelled pixels remains accurate. A candidate is a spectral band, described by its values on the small sample of labelled pixels, which come from two classes of land cover. It is characterised by $p=18$ signals belonging to four categories, all computed from the small sample: association with the class, redundancy, the band itself and the task. Eight signals measure how strongly the band is associated with the class. The absolute correlation captures a linear association, the mutual information any dependence, and the RELIEF score \citep{Ref4} how well each pixel is separated from its nearest neighbours of the other class. The absolute weights of the band in three linear classifiers, a logistic regression, an elastic-net logistic regression and a linear support vector machine, and its automatic relevance determination score \citep{Ref5} measure its usefulness given the other bands. The last of the eight detects bands that are informative only in combination with another band. These measures rank the bands differently from one scene to another, and none of them is reliable on its own when the sample is small, which is why the value model learns to combine them.

Three signals measure redundancy: the largest and the mean absolute correlation of the band with the other bands, and its correlation with the adjacent bands of the spectrum. Three signals describe the band itself: its variance, its position in the spectrum and its noise, measured as its deviation from the average of its two neighbouring bands. Four signals describe the task and are the same for all its bands: the logarithm of the ratio between the number of pixels and the number of bands, the logarithm of the number of bands, the balance between the two classes and the gain in accuracy of a nonlinear classifier over a linear one. They tell the value model which of the other signals to trust. The band-level signals are converted to ranks within the task, and the task-level signals keep their values.

The value is a binary indicator, equal to one when the band is among the 30 bands that have the highest mutual information with the class on a large labelled sample. It measures what the small sample cannot estimate reliably, how informative the band is on its own, and it does not depend on the other bands selected. Since the value is binary, the value model is a classifier, a multilayer perceptron with two hidden layers of 64 units whose output estimates the probability that the band is among the best ones.

The search is performed by simulated annealing with the diversity term of Equation~\eqref{eq:spread}, with the correlation distance $\delta(i,j)=1-|\rho_{ij}|$ as the dissimilarity, where $\rho_{ij}$ is the correlation between bands $i$ and $j$ on the small sample. Adjacent bands are often strongly correlated, so the bands with the highest predicted values tend to be neighbours that carry the same information. The weight of the diversity term is chosen for each task by cross-validation on the small sample, and all bands are selected in a single round. The hidden information is the class of the pixels outside the small sample, which determines the truly informative bands. The utility is the accuracy, on other pixels, of a logistic regression trained on the small sample restricted to the selected bands.

In input selection, a classifier is trained on tabular data with many input variables, and a few of them must be chosen so that the classifier remains accurate with fewer inputs. A candidate is an input variable, described by its values on the training samples. It is characterised by $p=15$ signals belonging to three categories, all computed from the training samples: association with the class, the variable itself and its redundancy, and the task. Eight signals measure how strongly the variable is associated with the class, with the same measures as for the bands: the absolute correlation, the mutual information, the RELIEF score, the absolute weights in three linear classifiers, the automatic relevance determination score, and a measure of the information the variable carries only in combination with another variable, as in an exclusive-or effect, which every marginal statistic misses. These measures rank the variables differently from one dataset to another, and a fixed choice of one of them can fail badly on a new dataset, which is why the value model learns to combine them.

Three signals describe the variable itself and its redundancy: its variance, and its largest and mean absolute correlation with the other variables. Four signals describe the task and are the same for all its variables: the logarithm of the ratio between the number of samples and the number of variables, the logarithm of the number of variables, the balance between the two classes, and the gain in accuracy of a nonlinear classifier over a linear one. The variable-level signals are converted to ranks within the task, and the task-level signals keep their values.

The value is a binary indicator, equal to one when the variable is informative about the class. It can be known exactly only on synthetic tasks, which are built with linear effects, interactions, exclusive-or effects or redundant copies of informative variables. Copies are labelled as not informative, since they add nothing to the variable they copy. The value model is a logistic regression, whose output estimates the probability that the variable is informative.

No diversity term is used. Redundancy enters the value only through the redundancy signals, and the variables with the highest predicted values are selected, in a single round. The hidden information is which variables are truly informative, which the training samples estimate only roughly. The utility is the accuracy, on held-out samples, of a logistic regression trained on the selected variables.

In experimental design, a design is a table that sets the level of each factor in each run of an experiment. Orthogonal minimally aliased response surface designs, or OMARS designs \citep{Ref6}, are three-level designs in which the main effects are uncorrelated with each other and with the second-order effects. Large designs of this type are obtained by enumerating and combining weighing designs \citep{ref50} or by a genetic algorithm \citep{ref69}. They can also be constructed by a simulated annealing that repeatedly swaps two levels within one column of the design, and a candidate is such a swap. Whether a swap improves the design can only be known by evaluating the quality criterion, which is the costly step. The signals are therefore chosen to be available without any evaluation. They are $p=12$ local quantities belonging to three categories: the state of the search, the swap and the column. Five describe the state of the search: the temperature, the fraction of the evaluation budget already used, the current value of the criterion, its distance to the best value met so far and the recent acceptance rate. Early in the search, at high temperature and far from a good design, many swaps improve the criterion, whereas very few do near the end. Five describe the swap itself: the position of the column, the two levels exchanged and the positions of the two rows. Two describe the column, the mean of its levels and the mean of their absolute values, which indicate how balanced the column is.

The value is a binary indicator, equal to one when the swap improves the criterion. It is recorded for every proposed swap along simulated annealing runs on designs of other sizes. A binary value is used because the decision to make is whether a swap is worth evaluating, not by how much it changes the criterion. The value model is a classifier, a multilayer perceptron or a logistic regression, chosen by the area under the receiver operating characteristic curve on held-out training designs.

The search is the simulated annealing itself, with SALVES used as a move filter, as described in Section~\ref{sec:salve}. At each step, four random swaps are proposed and only the swap with the highest predicted value is evaluated. The number of evaluations is therefore the same as without the value model, but it is spent on more useful moves. Each step selects a batch of one swap, so no diversity is needed. Nothing is hidden in the usual sense, since the criterion can be computed for any design, but each evaluation is costly. The effect of a swap on the criterion is known only once the swap has been evaluated, and this effect plays the role of the hidden information. The utility is minus the quality criterion, the mean absolute correlation between main effects and between main effects and second-order effects, plus 0.3 times the largest of these correlations. This criterion measures how far the design is from the orthogonality that defines an OMARS design.
\subsection{Experimental protocol}
\label{sec:protocol}

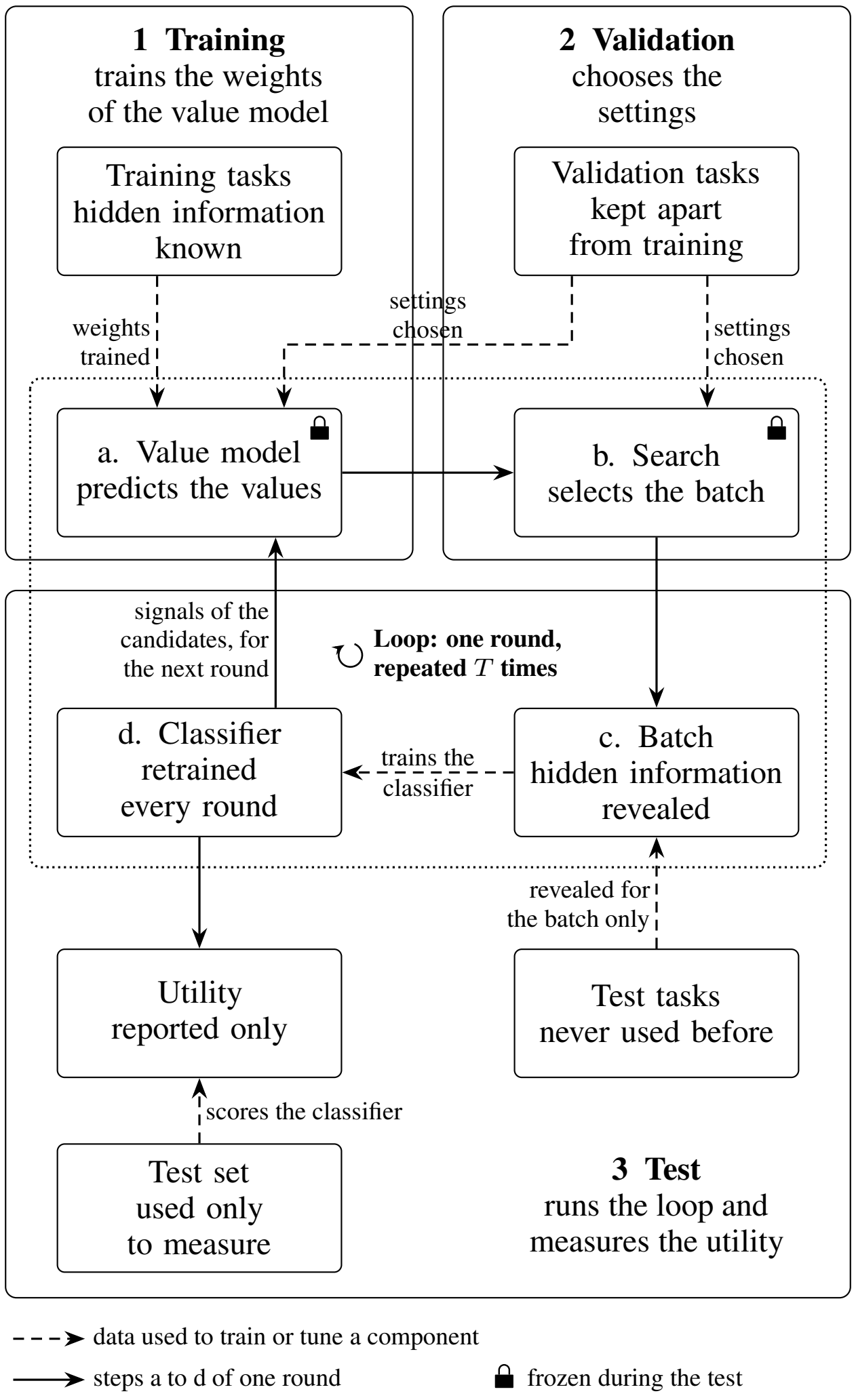
\begin{figure} [!h]
    \centering
\begin{tikzpicture}[
  font=\footnotesize,
  box/.style={draw, rounded corners=2pt, align=center, inner sep=3pt,
              text width=31*\ux, minimum height=12.5mm},
  arrow/.style={-{Stealth[length=2mm]}, semithick},
  data/.style={arrow, densely dashed},
  frame/.style={draw, rounded corners=3pt},
  loopframe/.style={draw, densely dotted, line width=0.6pt,
                    rounded corners=3pt},
  ptitle/.style={align=center, inner sep=0pt, anchor=north},
  lbl/.style={font=\scriptsize, inner sep=1.5pt, align=#1},
  lbl/.default=center,
  lock/.pic={
    \fill[rounded corners=0.2pt] (-0.9mm,-0.8mm) rectangle (0.9mm,0.5mm);
    \draw[line width=0.45pt] (-0.5mm,0.5mm) -- (-0.5mm,0.95mm)
      arc[start angle=180, end angle=0, radius=0.5mm] -- (0.5mm,0.5mm);},
  cycle/.pic={
    \draw[line width=0.5pt, -{Stealth[length=1.2mm]}]
      (70:1.15mm) arc[start angle=70, end angle=-240, radius=1.15mm];},
]
\pgfmathsetlengthmacro{\figw}{0.5*\textwidth}
\pgfmathsetlengthmacro{\ux}{\figw/100}

\begin{scope}[x=\ux, y=1mm]
\draw[frame] (-49.9,0)   rectangle (-1.8,-54);
\draw[frame] (1.8,0)     rectangle (49.9,-54);
\draw[frame] (-49.9,-57) rectangle (49.9,-126);

\node[ptitle] at (-25.9,-2)
  {\textbf{1\enspace Training}\\trains the weights\\of the value model};
\node[ptitle] at (25.9,-2)
  {\textbf{2\enspace Validation}\\chooses the\\ settings};

\draw[loopframe] (-47,-36.3) rectangle (47,-84);

\node[box] (train) at (-27,-20)     {Training tasks\\hidden information\\known};
\node[box] (valid) at ( 27,-20)     {Validation tasks\\kept apart\\from training};
\node[box] (vm)    at (-27,-45.5)   {a. Value model\\predicts the values};
\node[box] (sa)    at ( 27,-45.5)   {b. Search\\selects the batch};
\node[box] (clf)   at (-27,-74.75)  {d. Classifier\\retrained\\every round};
\node[box] (batch) at ( 27,-74.75)  {c. Batch\\hidden information\\revealed};
\node[box] (util)  at (-27,-98.25)  {Utility\\reported only};
\node[box] (pool)  at ( 27,-98.25)  {Test tasks\\never used before};
\node[box] (test)  at (-27,-117.25) {Test set\\used only\\to measure};
\node[align=center, inner sep=0pt] at (27,-117.25)
  {\textbf{3\enspace Test}\\runs the loop and\\measures the utility};

\pic at ([shift={(-2.2mm,-2.3mm)}]vm.north east) {lock};
\pic at ([shift={(-2.2mm,-2.3mm)}]sa.north east) {lock};

\draw[data] (train.south -| -32,0) -- (vm.north -| -32,0)
  node[lbl=right, midway, left] {weights\\trained};
\draw[data] (valid.south -| 17,0) -- (17,-33) -- (-17,-33)
  node[lbl, midway, above] {settings\\chosen}
  -- (vm.north -| -17,0);
\draw[data] (valid.south -| 33,0) -- (sa.north -| 33,0)
  node[lbl=left, midway, right] {settings\\chosen};
\draw[data] (batch.west) -- (clf.east)
  node[lbl, midway, above] {trains the}
  node[lbl, midway, below] {classifier};
\draw[data] (pool.north) -- (batch.south)
  node[lbl=right, pos=0.38, left] {revealed for\\the batch only};
\draw[data] (test.north) -- (util.south)
  node[lbl=left, midway, right] {scores the classifier};

\draw[arrow] (vm.east) -- (sa.west);
\draw[arrow] (sa.south) -- (batch.north);
\draw[arrow] (clf.north -| -18,0) -- (vm.south -| -18,0)
  node[lbl=right, pos=0.4, left] {signals of the\\candidates, for\\the next round};
\draw[arrow] (clf.south) -- (util.north);

\node[font=\scriptsize\bfseries, align=left, inner sep=0pt, anchor=west]
  (looplbl) at (-6.5,-63.3) {Loop: one round,\\repeated $T$ times};
\pic at ([xshift=-2.2mm]looplbl.west) {cycle};

\begin{scope}[font=\scriptsize]
\draw[data]  (-49,-130) -- ++(7mm,0)
  node[right, inner sep=2pt] {data used to train or tune a component};
\draw[arrow] (-49,-134) -- ++(7mm,0)
  node[right, inner sep=2pt] {steps a to d of one round};
\pic at (9,-134) {lock};
\node[right, inner sep=2pt] at (10.8,-134) {frozen during the test};
\end{scope}
\end{scope}
\end{tikzpicture}
    \caption{Roles of the data.}
    \label{fig:protocol}
\end{figure}

The experiments are designed so that the comparison between SALVES and other selection methods reflects what can be expected on a new problem. Three rules follow from this aim. Every choice made to build SALVES, from the parameters of the value model to the settings of the search, is made on data that play no part in the comparison. Every method is compared under the same conditions, with the same tasks, initial sets, budgets and evaluation procedure. The utility is measured on data that no method uses to select. Figure~\ref{fig:protocol} shows how the data are divided to satisfy these rules.

The data of each application are divided into three groups of tasks, which correspond to the three panels of Figure~\ref{fig:protocol}, and a test set used only to measure the utility. The training tasks provide the states and the true values on which the value model is fitted, as described in Section~\ref{sec:salve}. The validation tasks are kept apart from the training tasks and serve to choose the settings. On a validation task, the complete selection is run and its utility is measured on a validation set, which is distinct from the reference set used to compute the true values, so that the settings are not chosen on the data that define what the value model learns. The test tasks provide the problems on which SALVES and the baselines are compared. They come from a part of the data that is used neither to train the value model nor to choose the settings. The test set contains the held-out data on which the utility of the selected candidates is measured, and no method has access to it during the selection.

What a task is depends on the application. In active learning, a task is a pool of unlabelled images from a set of classes, together with a labelled test set from the same classes, and the images used to build training tasks are kept apart from those used to build test tasks. In band selection, a task is a pair of land-cover classes in one scene, with a small sample of labelled pixels and other labelled pixels on which the utility is measured, and training and test tasks come from different scenes. In input selection, the training tasks are synthetic datasets, generated with each of the four kinds of effects described in Section~\ref{sec:applications} and with varying numbers of samples, variables and informative variables, and the test tasks are real datasets. In experimental design, a task is the construction of a design with a given number of runs and factors, and the training tasks have other sizes than the test tasks.

All settings are chosen before any test task is used, and they are then frozen. The settings of the value model, such as its architecture, its learning rate and the signals it receives, are chosen by the measures of Section~\ref{sec:salve} on held-out training tasks, and by the utility of the complete selection on the validation tasks. The settings of the search, such as the number of groups, the weight of the typicality in the score, the weight of the diversity term and the point where the selection is handed over to margin sampling, are chosen by the utility on the validation tasks. When many settings may interact, they are screened with a definitive screening design \citep{Ref7}, a designed experiment that estimates the effect of each setting, and detects strong interactions, with few runs of the complete selection. The classifier of the evaluation procedure is the same for all methods, and its own settings are fixed on the training data before the comparison.

When the true values can be computed on the test tasks, the selection is also run with these true values, computed from the hidden information of the test tasks, in place of the predicted values. This run is not a selection rule, since it uses information that is unavailable in practice, but it shows how much utility the same search reaches when the value is known. The gap between SALVES and this reference measures what a better value model could add. It is a reference rather than an upper bound, since the true value is itself an approximation in some applications and the search remains the same.

Each application is compared with the selection methods that are standard for its problem.
\begin{itemize}
  \item In active learning, the baselines are random selection and four kinds of methods. Three uncertainty methods select the images with the smallest margin, the highest entropy or the lowest largest probability. Two methods favour representative images, $k$-center selection \citep{Ref8} and TypiClust \citep{Ref9}. A coverage method, ProbCover \citep{Ref10}, selects images that cover the pool. Two methods combine uncertainty and diversity, BADGE \citep{Ref11} and Cluster-Margin \citep{Ref12}.
  \item In band selection, the baselines are random bands, evenly spaced bands, the bands with the highest mutual information on the small sample, and two unsupervised methods, OPBS \citep{Ref13} and FDPC \citep{Ref14}.
  \item In input selection, each association signal used alone is a baseline. The other baselines are recursive feature elimination with a support vector machine \citep{Ref15}, mRMR \citep{Ref16} and L0Learn \citep{ref65, ref68}, a best-subset method from mixed-integer optimisation.
  \item In experimental design, the baseline is the simulated annealing that proposes swaps at random, with the same number of evaluations.
\end{itemize}
All baselines receive the same tasks, initial sets, budgets, classifier and seeds as SALVES, and their settings follow their original publications. \textcolor{black}{General methods for subset selection, such as the greedy maximisation of a coverage function, require a set function that can be evaluated when the selection is made. In the problems considered here, the utility is hidden at that time, so these methods can only be applied to a function computed from the available information. This is what several baselines do: $k$-center and ProbCover greedily cover the pool, and mRMR greedily maximises a criterion based on mutual information.}

The seeds fix the task, the initial set and the randomness of the classifier, and every method is run with the same seeds, so that the methods are compared seed by seed. For a given seed, the difference in utility between SALVES and a baseline removes the part of the variation that comes from the task. \textcolor{black}{The results report the mean of these differences with its 95\% confidence interval. A difference whose interval includes zero is reported as tied.} When the selection has several rounds, the utility is reported after each round and as its mean over rounds, the first round excluded, since the initial set is the same for all methods. The value model is trained once and shared by all seeds, as in a practical use of the framework. The computation time of the learning phase and of the selection at each round is also measured, for SALVES and for the baselines.

A value model is useful in practice only if it can be reused on tasks that differ from those on which it was trained. In every application, the test tasks already differ from the training tasks, since they come from other pools, other scenes, other datasets or designs of other sizes. Transfer tests measure how far this reuse extends, by making the test tasks differ in one more property, while the value model and the settings are kept exactly as they were frozen. In active learning, the test tasks may contain classes that never appear in the training tasks, the images may be described by a different pretrained network, or the test tasks may come from another dataset. In band selection, the scenes may come from another sensor. Each transfer test is compared with a value model trained on tasks of the same kind as the test tasks, whenever such tasks are available. The difference between the two measures what is lost by reusing the value model instead of training it again.

The selection is made with one classifier, but the selected candidates may later be used with another. The utility is therefore also measured with classifiers that play no part in the selection, trained on the same selected candidates. In active learning, these are a classifier that assigns each image to the class with the nearest mean, a $k$-nearest-neighbour classifier, a multilayer perceptron and label propagation \citep{Ref17}, which also uses the unlabelled images of the pool. A version of the classifier that weights each class by the inverse of its frequency in the selected set tests whether an unbalanced selection of classes explains the results. In band selection, a one-dimensional convolutional network replaces the logistic regression, and in input selection, the logistic regression is replaced by support vector machines with a linear and a radial basis kernel. A selection that helps only the classifier it was made for would be of limited use, and these measurements test whether the gain carries over.

\subsection{Ablations}
\label{sec:ablations}
The ablations are designed to identify which components of SALVES contribute to its performance. Each ablation modifies one component while keeping all other components fixed at the settings selected on the validation tasks. The remaining parts of SALVES are not retuned, so that differences can be attributed to the component being studied rather than to additional tuning. Each ablation is evaluated on the test tasks using the same seeds as SALVES and is compared with SALVES through the paired differences described in Section~\ref{sec:protocol}. The components are identified by their steps in Figure~\ref{fig:overview}, and an ablation is performed in every application where the corresponding component is present.

The first group of ablations examines the value used in steps 3, 5 and 8 of Figure~\ref{fig:overview}. The main question is whether learning the value provides an advantage over simpler ways of ranking candidates. Within the same search procedure, the predicted values are therefore replaced by random scores or by hand-crafted scores based on the margin or the typicality of each candidate. In band selection and input selection, this comparison is complemented by baselines that rank candidates using a single association signal. \textcolor{black}{In input selection, the value model is also replaced by equal weights of the same signals.} The true-value run described in Section~\ref{sec:protocol} provides the other end of the comparison. These experiments are followed by modifications of the value model itself while keeping its role unchanged. Candidate-level models that process one candidate at a time include linear models, kernel models and multilayer perceptrons of different sizes. Models that process the candidates jointly include a set transformer, a graph network and a convolutional network that operates directly on the image. In active learning, the effect of the input signals is examined by comparing the seven selected signals with larger sets containing up to 38 signals, added family by family. Finally, reducing the number of training tasks and the number of candidates per state assesses how much training data is required to learn useful values.

The next group of ablations examines the description of the candidates, which enters the state in steps 2 and 6. This matters most in active learning, where each image is described using features from a pretrained network. The choice of this network affects the classifier, the distances between images, and consequently several of the signals used by SALVES. Replacing the pretrained network with networks of different sizes and architectures therefore tests the sensitivity of the method to the candidate representation. Two variants distinguish between the effect of changing the representation itself and the effect of changing the representation on what the value model has learned. In one, the value model is retrained using the new description, testing whether the framework can operate with different image representations. In the other, the value model trained with the original description is retained, testing whether the learned values transfer across representations. The latter corresponds to the transfer test described in Section~\ref{sec:protocol}, with the pretrained network as the property that changes.

The search in step 9 is examined by varying both the diversity mechanism and the optimisation procedure. Removing diversity entirely makes the batch consist of the $B_t$ candidates with the highest predicted values and therefore isolates the contribution of diversity. The diversity mechanisms are then varied within their respective forms. Under the group constraint, the number of groups per candidate in the batch and the weight assigned to typicality are varied, and an additional term rewarding coverage of the pool is introduced. For the diversity term, the spread of Equation~\eqref{eq:spread} is compared with the quadratic R\'enyi entropy of the selected images, minus the logarithm of the average Gaussian similarity between all pairs of selected images, a common measure of set diversity. The role of the search is assessed separately. With the diversity term, simulated annealing is replaced by a greedy search that adds, at each step, the candidate that gives the largest increase in the objective. Under the group constraint, where the exact batch of Section~\ref{sec:salve} is the default, this batch is compared with simulated annealing started from it. In experimental design, the number of swaps scored at each step is also varied, controlling the balance between guidance from the value model and randomness in the search.

The final group of ablations concerns the large-budget rule used in active learning. SALVES hands the selection over to margin sampling once enough labels have been selected. Removing this rule allows SALVES to select candidates until the budget is exhausted, while moving the handover point tests the sensitivity to when the transition occurs. A further variant replaces margin sampling at the handover point with the candidates having the highest predicted values. This comparison determines whether the learned value remains informative at large budgets once the diversity mechanism is no longer needed, or whether the classifier's margin provides a better guide at that stage.
{\color{black}\section{Results}
\label{sec:results}
\subsection{Experimental Setup}
\label{sec:res-setup}
The framework of Section~\ref{sec:salve} is evaluated on four selection problems, batch active learning, band selection, input selection and experimental design. These problems come from different communities and differ in the nature of their candidates, in the number of candidates and in the information available to select them. Two questions guide this evaluation. The first is whether a single framework can compete with the methods designed specifically for each problem. The second is whether what is learned on the training tasks remains useful on test tasks that differ from them. Sections~\ref{sec:res-al} to~\ref{sec:res-design} report the results of the four applications. Section~\ref{sec:res-analyses} then analyses the framework as a whole: the contribution of each of its components, the quality of its search and its computational cost.

The four problems differ in both the type of object selected and the number of candidates. Over the training and test tasks, the number of candidates ranges from 2{,}000 to 640{,}000 images in active learning, from 103 to 204 spectral bands in band selection, from 50 to 10{,}000 variables in input selection and from 1{,}300 to 36{,}800 possible swaps at each step in experimental design. In all four applications, the test tasks are kept apart from the tasks on which the value model is trained. The distinction between training and test tasks is based on the images or the pair of digits in active learning, on the sensor in band selection, on the origin of the data in input selection and on the size of the design in experimental design. The following subsections first describe the data used in each application, along with the number of seeds and the budgets.

All differences are paired over seeds, as described in Section~\ref{sec:protocol}, and are defined so that positive values indicate better performance of the SALVES framework. A run denotes one seed on one test task. Each comparison is made against the best baseline, defined as the baseline with the highest mean utility on the test tasks. This baseline is identified retrospectively and would therefore not be known to a user in advance. Two methods are said to be tied when the 95\% interval of their paired difference includes zero, and intervals are computed over runs unless stated otherwise.

\subsection{Batch active learning}\label{sec:res-al}
Active learning is examined on ImageNet \citep{ref46}, using the version with blurred faces \citep{pmlr-v162-yang22q}, and on pairs of MNIST digits \citep{ref47}. For ImageNet, each image is represented by features extracted using DINOv2 ViT-S/14 \citep{ref43}, a frozen network pretrained without labels, and classified using logistic regression. Within each class, the images are split into two halves: the first is used to construct the training and validation tasks, while the second is reserved for the test pools. Utility is measured on the 50{,}000 images of the official ImageNet validation set, which serves as the test set and is not used to choose the settings. The training tasks provide 72 states, one per round of a random selection. The value model is fitted on 54 states, using the true values of 10{,}000 candidates per state, while the remaining 18 states are reserved for validation. The budget is expressed as the average number of labelled images per class, or labels per class for short, since only the initial set, containing one image per class, is balanced. The evaluation uses three pools, all scored on the same balanced test set.
\begin{itemize}
\item The reduced pool contains 100 images per class. It covers very low budgets, from 1 to 10 labels per class, with 20 seeds.
\item The full pool contains about 640{,}000 images. It extends the comparison to larger budgets, up to 400 labels per class, with five seeds.
\item The long-tailed pool contains about 129{,}000 images, with classes ranging from 600 images for the largest down to 6 for the smallest, and its initial set of 1{,}000 images is drawn at random. It makes the task harder and is used in the transfer tests, from 1 to 10 labels per class, with ten seeds.
\end{itemize}

For MNIST, the evaluation considers nine binary classification tasks, each corresponding to a pair of digits and using a pool of 2{,}000 images. The classifier, also a logistic regression, is trained directly on the pixel values. The value model is fitted on 27{,}000 examples drawn from 30 other digit pairs. Each task is evaluated with budgets ranging from 10 to 80 labels and ten seeds.

Unlike ImageNet, the true value can be computed exactly as the change in accuracy after retraining the classifier. Moreover, the batches are small, so the search uses the diversity term instead of the group constraint, and the selection is never handed over to margin sampling. Finally, since there are only two classes, entropy and the lowest largest probability induce the same selection as margin sampling. Evaluating them separately would therefore be redundant. ProbCover and Cluster-Margin are not evaluated on these tasks.

\begin{figure}[!h]
\centering
\includegraphics[width=\textwidth]{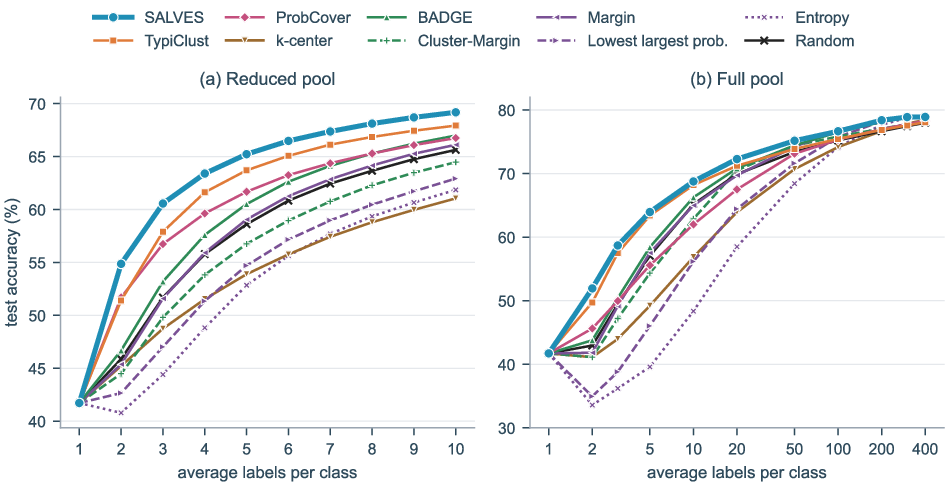}
\caption{Active learning on ImageNet.}
\label{fig:res-al-imagenet}
\end{figure}

Figure~\ref{fig:res-al-imagenet} shows the comparison of SALVES with the different baselines on ImageNet. On the reduced pool (panel a), SALVES outperforms every baseline at all budgets from two labels per class onwards. Averaged over these budgets, its accuracy exceeds that of TypiClust, the strongest baseline, by 1.76 percentage points (Table~\ref{tab:al-reduced}). The lead is largest in the most data-scarce setting, exceeding 3 points over ProbCover with only two labels per class, before stabilising at approximately 1.25 points over TypiClust from seven labels per class onwards. TypiClust is the strongest baseline from three labels per class, followed by ProbCover up to eight and by BADGE at nine and ten. By contrast, four other baselines, $k$-center, Cluster-Margin, entropy and the lowest largest probability, perform below random selection at every budget. For entropy and the lowest largest probability, this is consistent with the known unreliability of uncertainty estimates when very few labels are available \citep{Ref9}. The advantage does not depend on the network that extracts the features. The comparison is repeated with two larger networks, DINOv2 ViT-B/14 and ViT-L/14, and with a weaker one, a ResNet-50 pretrained with DINO \citep{ref41}, with a new value model trained on the features of each. The accuracy of every method rises or falls with the network, but SALVES remains 1.3 to 1.7 points ahead of TypiClust.
\begin{table}[!h]
\centering
\small
\setlength{\tabcolsep}{4pt}
\caption{Reduced pool of ImageNet.}
\label{tab:al-reduced}
\begin{tabular}{lrrll}
\hline
Method & Accuracy & Difference & 95\% interval & SALVES \\
\hline
SALVES & 64.88 & & & \\
TypiClust & 63.11 & 1.76 & [1.66, 1.86] & Ahead \\
Class-balanced random & 62.41 & 2.46 & [2.33, 2.60] & Ahead \\
ProbCover & 61.72 & 3.16 & [3.05, 3.27] & Ahead \\
BADGE & 60.36 & 4.52 & [4.40, 4.63] & Ahead \\
Margin & 59.06 & 5.81 & [5.69, 5.93] & Ahead \\
Random & 58.81 & 6.07 & [5.96, 6.18] & Ahead \\
Cluster-Margin & 57.20 & 7.68 & [7.52, 7.83] & Ahead \\
Lowest largest prob. & 55.22 & 9.66 & [9.51, 9.81] & Ahead \\
$k$-center & 54.71 & 10.17 & [9.99, 10.34] & Ahead \\
Entropy & 53.56 & 11.31 & [11.12, 11.51] & Ahead \\
\hline
\end{tabular}
\end{table}

Two additional checks determine whether this gain is attributable to class balance rather than to the choice of images. First, random selection with the same number of images in every class, which uses the labels of the pool and is therefore not available in practice, stays 2.5 points below SALVES (Table~\ref{tab:al-reduced}). Second, when the classifier is trained with class weights, which compensate for an unbalanced labelled set, SALVES gains 0.4 points and remains ahead of every baseline, by 1.6 points over TypiClust. Selecting the classes in equal proportions therefore does not explain the gain.

The full pool shows how long this lead lasts as the budget grows (Figure~\ref{fig:res-al-imagenet}b). Up to 50 labels per class, SALVES is ahead of the best baseline at every budget except five labels per class, where the two are tied. Over this range, it is 1.14 points ahead of TypiClust, and the lead does not fall steadily, since it is larger at 20 labels per class than at 10. Beyond 50 labels per class, the selection is handed over to margin sampling and the curves come together. SALVES falls 0.17 points below the best baseline at 100 labels per class and is tied with it from 200. At these budgets, SALVES and margin sampling select with the same rule, so the gap between them comes only from the images each had labelled up to 50 labels per class. With the ResNet-50 pretrained with DINO, SALVES is ahead up to 50 labels per class by smaller amounts, slightly below from 100 to 300 and tied at 400 (Table~\ref{tab:al-whole}). The lead of SALVES is therefore confined to the budgets where labels are scarce.

\begin{table}[!h]
\centering
\small
\setlength{\tabcolsep}{4pt}
\caption{Full pool of ImageNet.}
\label{tab:al-whole}
\begin{tabular}{rlrllrl}
\hline
 & \multicolumn{3}{c}{ViT-S/14} & \multicolumn{3}{c}{ResNet-50 (DINO)} \\
Labels per class & Best baseline & Difference & SALVES & Best baseline & Difference & SALVES \\
\hline
2 & TypiClust & 2.21 & Ahead & TypiClust & 1.57 & Ahead \\
3 & TypiClust & 1.20 & Ahead & TypiClust & 0.86 & Ahead \\
5 & TypiClust & 0.56 & Tied & TypiClust & 0.55 & Ahead \\
10 & TypiClust & 0.56 & Ahead & TypiClust & 0.42 & Ahead \\
20 & TypiClust & 1.05 & Ahead & Cluster-Margin & 0.80 & Ahead \\
50 & BADGE & 0.72 & Ahead & Cluster-Margin & 0.39 & Ahead \\
100 & Margin & $-0.17$ & Behind & BADGE & $-0.35$ & Behind \\
200 & Margin & 0.02 & Tied & Margin & $-0.24$ & Behind \\
300 & Lowest largest prob. & 0.01 & Tied & Margin & $-0.12$ & Behind \\
400 & BADGE & $-0.02$ & Tied & Margin & $-0.05$ & Tied \\
\hline
\end{tabular}
\end{table}

On the digit pairs, as presented in Figure~\ref{fig:res-al-digits}a, SALVES is 0.46 points above BADGE, the strongest baseline, across budgets of 20 to 80 labels. The difference is positive on all nine pairs, but it is uneven. It reaches 2.10 points on the pair of digits 4 and 9, against 0.26 points on average on the eight other pairs. The lead is therefore clear over the 90 runs, but SALVES and BADGE are tied when each pair counts once.

\begin{figure}[!h]
\centering
\includegraphics[width=\textwidth]{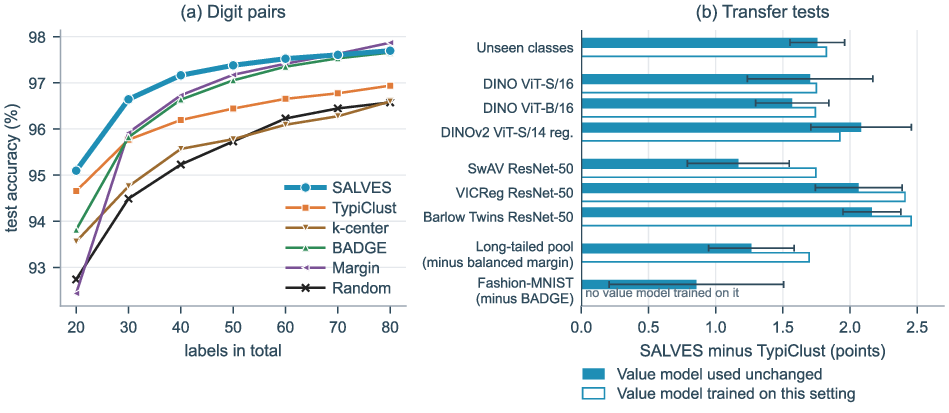}
\caption{Digit pairs and transfer tests.}
\label{fig:res-al-digits}
\end{figure}

The curves, which start at 20 labels after the first selected batch, differ from those of ImageNet (Figure~\ref{fig:res-al-digits}a). Margin sampling starts as the lowest of all methods at 20 labels, but it is tied with BADGE from 30 to 70 labels, so BADGE is the best baseline over the whole range mainly because of its better start. From 30 labels onwards, the best baseline at each budget is margin sampling. SALVES is ahead of it up to 50 labels, by 0.72 points at 30, the two are tied at 60 and 70, and margin sampling is ahead by 0.17 points at 80. The value model was trained on states with at most 30 labels and the selection is never handed over to margin sampling, so these curves show SALVES itself beyond the range of its training states.

Figure~\ref{fig:res-al-digits}b gathers the transfer tests, in which the value model is used unchanged on tasks that differ from its training tasks. Table~\ref{tab:al-transfer} also gives the cost of transfer, the difference between this value model and one trained on the test setting. When the test tasks contain only classes 500 to 999 and the value model is trained on classes 0 to 499, SALVES is 1.76 points above TypiClust, and a value model trained on classes 500 to 999 is tied with the transferred one. When the features come from six further networks \citep{ref41,ref44,ref39,ref42,ref40} and the value model trained on ViT-S/14 is used unchanged, SALVES is 1.17 to 2.16 points above TypiClust, ahead on every network. Compared with a value model trained on the same network, the transferred one is tied on two of the three transformers and ahead by 0.16 points on the third, and it is behind by 0.29 to 0.58 points on the three ResNet-50 networks. Changing the architecture therefore seems to cost more than changing the pretraining method, although three networks of each kind give only an indication. On the long-tailed pool, the best baseline is a margin sampling that takes the predicted classes in turn, close to the method of \citet{ref67} and called balanced margin sampling here. Over 2 to 10 labels per class, SALVES with the transferred value model is 1.27 points above it, and a value model trained on long-tailed pools adds 0.43 points, although balanced margin sampling overtakes SALVES from eight labels per class. Finally, the value model trained on digits is used unchanged on nine pairs of classes of Fashion-MNIST \citep{ref38}, with 20 to 40 labels. On these tasks, the strongest baseline is TypiClust. SALVES is 0.40 points above it, a difference reported as tied. It outperforms BADGE by 0.86 points, as shown in the figure, close to the 0.88 points obtained on the digit pairs at the same budgets. In this run, the signal that gives the fraction of rounds completed was left at zero, which changes the result on the digit pairs at these budgets by 0.03 points.

\begin{table}[!h]
\centering
\small
\setlength{\tabcolsep}{4pt}
\caption{Transfer of the value model.}
\label{tab:al-transfer}
\begin{tabular}{lrlrrl}
\hline
 & \multicolumn{2}{c}{Transferred} & Trained on it & \multicolumn{2}{c}{Cost of transfer} \\
Test setting & Difference & Result & Difference & Difference & Result \\
\hline
Unseen classes & 1.76 & Ahead & 1.82 & $-0.07$ & Tied \\
DINO ViT-S/16 & 1.70 & Ahead & 1.75 & $-0.05$ & Tied \\
DINO ViT-B/16 & 1.57 & Ahead & 1.74 & $-0.17$ & Tied \\
DINOv2 ViT-S/14 reg. & 2.08 & Ahead & 1.92 & 0.16 & Ahead \\
SwAV ResNet-50 & 1.17 & Ahead & 1.75 & $-0.58$ & Behind \\
VICReg ResNet-50 & 2.06 & Ahead & 2.41 & $-0.34$ & Behind \\
Barlow Twins ResNet-50 & 2.16 & Ahead & 2.45 & $-0.29$ & Behind \\
Fashion-MNIST (minus BADGE) & 0.86 & Ahead & -- & -- & -- \\
Long-tailed pool & 1.27 & Ahead & 1.70 & $-0.43$ & Behind \\
\hline
\end{tabular}
\end{table}

The gain also holds with classifiers that play no part in the selection. When other classifiers are trained on the selected images, on the reduced pool with ten seeds, the difference with TypiClust is 1.84 points with the logistic regression used in the selection and with a multilayer perceptron, and it falls to 0.88 points with the nearest class mean, 0.89 with the $k$-nearest-neighbour classifier and 0.75 with label propagation. SALVES therefore remains ahead with every classifier, although the gain is about half as large with the three classifiers that differ most from the logistic regression that defines the true values.

Batch active learning therefore shows both what SALVES brings and where it stops. When labels are scarce, SALVES is ahead of the best baseline at almost every budget, up to 50 labels per class on ImageNet and up to 50 labels in total on the digit pairs. The gain is largest at the lowest budgets, where it exceeds 3 points on the reduced pool. This result holds with the four networks used to provide the features, with a classifier trained with class weights and with classifiers that play no part in the selection, although the gain is about half as large with the classifiers furthest from the logistic regression. The value model also carries over to new tasks. Used unchanged on other classes and on six other networks, it keeps SALVES 1.2 to 2.2 points above TypiClust, on a long-tailed pool 1.3 points above the best baseline, and on Fashion-MNIST 0.9 points above BADGE, while training it again on the test setting adds at most 0.6 points. The gain shrinks as the budget grows, and margin sampling catches up at about 100 labels per class on ImageNet and at about 60 labels on the digit pairs. SALVES is therefore best suited to small labelling budgets, where uncertainty is unreliable, and margin sampling remains the better choice once the classifier is accurate.
\subsection{Band selection}\label{sec:res-bands}
Band selection is examined on hyperspectral images from three scenes \citep{webpage}. Each scene is an area of land imaged by an airborne sensor, in which every pixel has a spectrum, with one value per band, and the labelled pixels have a land cover class. Two scenes, Indian Pines and Salinas, provide the training tasks. They were acquired with the same sensor and contain 200 and 204 spectral bands respectively. In each scene, 30 pairs of land cover classes are drawn, and each pair defines two training tasks, which differ in their labelled pixels. The 120 training tasks contain 24{,}240 bands whose true value is known. The value model is a multilayer perceptron. It was preferred to three linear models on validation pairs, where its area under the ROC curve is 0.85.

The test tasks come from the third scene, Pavia University, which was acquired with a different sensor and contains 103 bands. The value model is used unchanged on this scene, so that this setup evaluates the transfer to a different scene and a different sensor. Such a transfer is possible because the signals of each band are converted to ranks within the task, which makes them comparable from one scene to another.

Each test task separates two of the nine Pavia University classes from 100 labelled pixels, 50 per class. There are 20 pairs of classes, drawn at random, and ten seeds per pair. The number of bands to select takes 11 values, ranging from 3 to 20, which makes 200 runs for each of them. The utility is measured by the accuracy of a logistic regression trained on the 100 labelled pixels and tested on up to 1{,}600 other pixels. As a reference, a logistic regression trained on all 103 bands achieves an accuracy of 94.11\%.

SALVES is compared with five baselines. Random band selection and evenly spaced band selection use no data. OPBS and FDPC use the spectra of the 100 labelled pixels but do not use their labels. The fifth baseline selects the bands with the highest mutual information with the class on these pixels. For SALVES, the weight of the diversity term is chosen in each run by cross-validation on the same 100 pixels. The runs of one pair share the same two classes and are therefore not independent. For this reason, the intervals of this subsection are computed over the 20 pairs. Finally, the comparison is repeated at four and at ten bands with a one-dimensional convolutional network in place of the logistic regression.

\begin{figure}[!h]
\centering
\includegraphics[width=\textwidth]{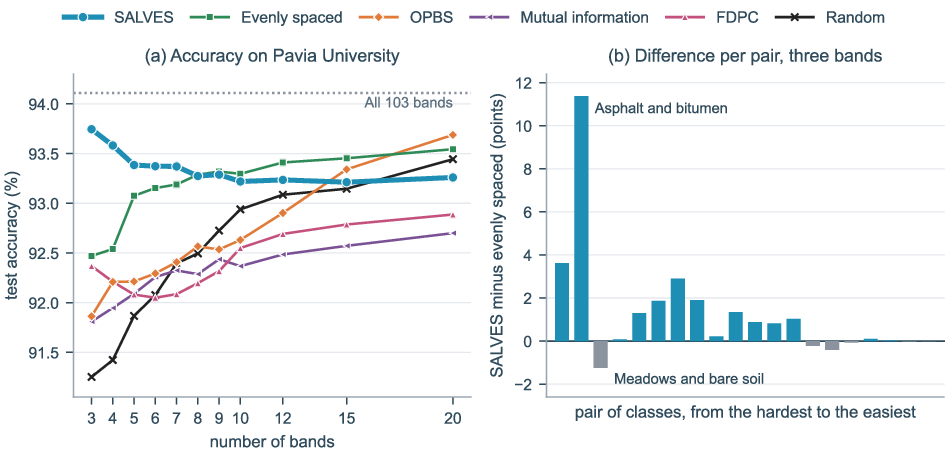}
\caption{Band selection results}
\label{fig:res-bands}
\end{figure}

Figure~\ref{fig:res-bands} compares SALVES with the five baselines. SALVES performs best when the selection budget is small, which is where the choice of the bands matters most. With three bands, it is ahead of each of the five baselines, by 1.28 points over evenly spaced bands, the best of them, and by 1.4 to 2.5 points over the others. With four bands, it is again ahead of all five, by 1.04 points over evenly spaced bands. Three of the 103 bands are enough for SALVES to come within 0.4 points of the accuracy obtained with all bands. No baseline does better with any number of bands up to twenty: the best of them, OPBS with twenty bands, reaches the accuracy that SALVES obtains with three. Table~\ref{tab:res-bands} gives the accuracy of every method, with the best baseline in bold, and the difference between SALVES and this baseline.

\begin{table}[!h]
\centering
\small
\setlength{\tabcolsep}{4pt}
\caption{Accuracy (\%) on Pavia University.}
\label{tab:res-bands}
\begin{tabular}{rrrrrrrrl}
\hline
 & & Evenly & Mutual & & & & & \\
Bands & Random & spaced & information & OPBS & FDPC & SALVES & Difference & Result \\
\hline
3 & 91.25 & \textbf{92.47} & 91.81 & 91.86 & 92.37 & 93.74 & 1.28 & Ahead \\
4 & 91.42 & \textbf{92.54} & 91.95 & 92.21 & 92.21 & 93.58 & 1.04 & Ahead \\
5 & 91.87 & \textbf{93.08} & 92.09 & 92.21 & 92.08 & 93.38 & 0.31 & Tied \\
6 & 92.08 & \textbf{93.15} & 92.26 & 92.29 & 92.05 & 93.37 & 0.22 & Tied \\
7 & 92.39 & \textbf{93.19} & 92.32 & 92.41 & 92.09 & 93.37 & 0.18 & Tied \\
8 & 92.50 & \textbf{93.29} & 92.29 & 92.57 & 92.20 & 93.27 & $-0.01$ & Tied \\
9 & 92.72 & \textbf{93.32} & 92.44 & 92.53 & 92.32 & 93.29 & $-0.03$ & Tied \\
10 & 92.94 & \textbf{93.30} & 92.37 & 92.63 & 92.55 & 93.22 & $-0.08$ & Tied \\
12 & 93.09 & \textbf{93.41} & 92.48 & 92.90 & 92.69 & 93.23 & $-0.18$ & Tied \\
15 & 93.15 & \textbf{93.45} & 92.57 & 93.34 & 92.79 & 93.21 & $-0.24$ & Tied \\
20 & 93.44 & 93.54 & 92.70 & \textbf{93.69} & 92.89 & 93.26 & $-0.43$ & Tied \\
\hline
\end{tabular}
\end{table}

The gain is concentrated on class pairs that are difficult to separate. For seven of the 20 pairs, three evenly spaced bands already achieve an accuracy above 99\%, leaving little room for improvement. On each of these pairs, SALVES remains within 0.4 points of this baseline. On the 13 other pairs, SALVES is ahead on 12 and gains 2.0 points on average. The largest gain is on asphalt against bitumen, where three bands give 78.3\% instead of 67.0\%. The result does not rest on this pair alone. Without it, the difference with evenly spaced bands is still 0.74 points at three bands and 0.71 points at four, and both intervals exclude zero. The largest loss on a pair is 1.2 points, on meadows against bare soil.

With more bands, the accuracy of SALVES remains stable, between 93.2\% and 93.7\%, while the baselines need additional bands to approach it. Between three and twenty bands, evenly spaced bands gain 1.1 points, OPBS 1.8 points and random bands 2.2 points. From five to twenty bands, SALVES is tied with the best baseline at every number of bands. SALVES achieves a higher mean accuracy than evenly spaced bands for budgets of up to seven bands and remains within 0.3 points of it at larger budgets. With twenty bands, OPBS becomes the strongest baseline. Its accuracy is 0.43 points above that of SALVES, a difference that leaves the two methods tied.

Over the 11 budgets, SALVES is never behind any of the five baselines. It also outperforms mutual information at every budget, FDPC at seven of them and random selection at six. Averaged across the 11 budgets, SALVES achieves the highest accuracy among the six methods. Its mean accuracy is approximately one point higher than that of random selection, FDPC and mutual information, and tied with that of evenly spaced bands and OPBS. Section~\ref{sec:res-analyses} examines the roles of the value model and of the diversity term at three and at ten bands.

With the convolutional network, the choice of the bands matters much less. At four and at ten bands, the best baseline, FDPC, is only 0.2 points above random bands, and evenly spaced bands lie between the two. SALVES is tied with the best baseline at both numbers of bands, 0.16 and 0.57 points below FDPC. The gain of SALVES therefore appears where the choice of the bands matters, with few bands and a linear classifier.

To sum up, band selection shows that a value model trained on one sensor can be used unchanged on another. With three or four bands, SALVES selects better bands than each of the five baselines, and its three bands reach the accuracy that the best baseline obtains with twenty. The gain is concentrated on the difficult class pairs. With more bands, SALVES has the highest average accuracy of the six methods and is never behind a baseline, although it is tied with the best of them from five bands onwards. With a convolutional network, all selections are close to one another and SALVES is tied with the best baseline.
\subsection{Input selection}\label{sec:res-inputs}
Input selection is examined on five datasets with two classes. Four of them are biomedical datasets in which the number of variables exceeds the number of samples: Colon \citep{ref45} (62 samples, 2{,}000 variables), Leukemia \citep{ref53} (72 samples, 7{,}129 variables), Prostate \citep{ref49} (102 samples, 6{,}033 variables) and Arcene \citep{ref48} (200 samples, 10{,}000 variables). The fifth dataset, Madelon \citep{ref48}, is an artificial benchmark with 2{,}600 samples and 500 variables.

The value model is trained on synthetic tasks only, in which the variables that determine the class are known. There are 240 such tasks of four types, with 50 to 1{,}000 variables and 60 to 800 samples. The model is a logistic regression on the signals of Section~\ref{sec:applications}, which are converted to ranks within each task. It is used unchanged on the five real datasets, so that the whole evaluation is a transfer from synthetic to real data. The logistic regression was preferred to a multilayer perceptron after a comparison on Colon, Leukemia and Madelon. Arcene and Prostate were added afterwards and took no part in any choice.

In each run, a dataset is split at random into 70\% for training and 30\% for testing. The signals and the selection use the training part only. Three classifiers are then trained on the selected variables: a logistic regression, a linear support vector machine and a support vector machine with a Gaussian kernel. The number of selected variables takes nine values, from 5 to 100. The utility of a run is the test accuracy averaged over the three classifiers and the nine numbers of variables. There are ten seeds for Arcene and Prostate and 30 seeds for the three other datasets, which makes 110 runs.

SALVES selects the variables with the highest predicted value, without a separate diversity term. It is compared with six baselines. Four of them rank the variables with a fixed criterion: mutual information, mRMR, RELIEF and the absolute weights of a logistic regression trained on all variables. The fifth baseline, SVM-RFE, removes variables step by step with a linear support vector machine. \textcolor{black}{The sixth baseline, L0Learn \citep{ref65, ref68}, comes from mixed-integer optimisation and fits a logistic regression with a penalty on the number of variables.} Two references complete the comparison. The first is the classifiers trained on all variables. The second is the average of the eight association signals of the value model with equal weights, which shows what the learned weights add. The intervals are computed over runs, with each dataset given the same weight.

\begin{figure}[!h]
\centering
\includegraphics[width=\textwidth]{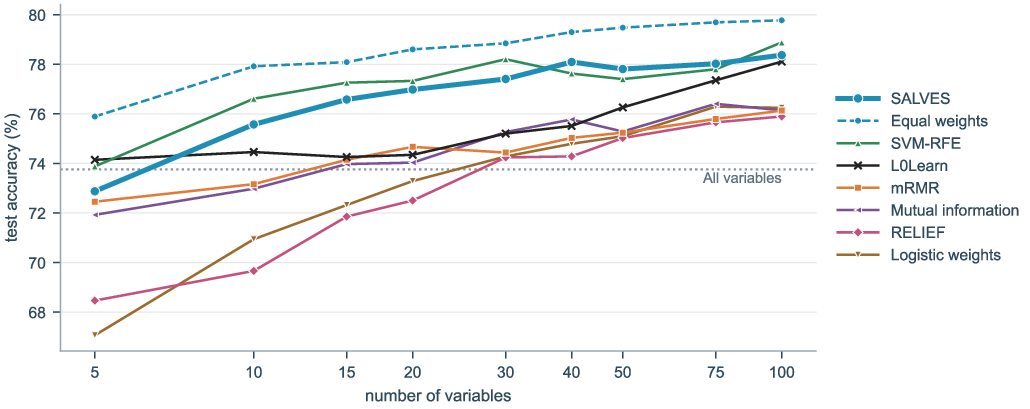}
\caption{Input selection on five datasets.}
\label{fig:res-inputs}
\end{figure}

Figure~\ref{fig:res-inputs} shows the accuracy of each method against the number of variables, averaged over the five datasets. SALVES and SVM-RFE stand apart from the other baselines. From 10 to 100 variables, SALVES is ahead of the best of the four ranking baselines at every number of variables, by 1.6 to 2.5 points. With SVM-RFE, the best baseline, it is tied at all nine numbers of variables. \textcolor{black}{SALVES outperforms L0Learn by 1.3 points on average. The gain is concentrated between 15 and 50 variables, and the two methods are tied at the other numbers of variables.} Over the whole range, SALVES is ahead of mutual information by 2.2 points, of mRMR by 2.3 points, of the logistic weights by 3.5 points and of RELIEF by 3.8 points. It is also 3.1 points above the classifiers trained on all variables.

The main advantage of SALVES is its reliability from one dataset to another, as presented in Table~\ref{tab:res-inputs}, where the best baseline of each dataset is in bold. Each of the four ranking baselines is the best on one or two datasets and fails on another. Mutual information is the best on Arcene and is 17 points below the best baseline on Prostate. The logistic weights are the best on Prostate and Madelon and are 18 points below it on Arcene. RELIEF is the best on Colon and mRMR on Leukemia, and each is 14 points below the best baseline on another dataset. A user cannot know in advance which criterion suits a new dataset. SALVES removes this choice. On every dataset it is within 3.8 points of the best baseline, as is SVM-RFE.

\begin{table}[t]
\centering
\small
\setlength{\tabcolsep}{4pt}
\caption{Accuracy (\%) on the five datasets.}
\label{tab:res-inputs}
\begin{tabular}{lrrrrrrrrr}
\hline
 & Mutual & & & Logistic & & & & Equal & All \\
Dataset & information & mRMR & RELIEF & weights & SVM-RFE & L0Learn & SALVES & weights & variables \\
\hline
Arcene & \textbf{74.94} & 73.65 & 60.67 & 56.77 & 71.14 & 70.34 & 73.65 & 74.49 & 80.06 \\
Prostate & 60.50 & 63.57 & 68.18 & \textbf{77.85} & 76.63 & 74.64 & 74.01 & 81.30 & 63.12 \\
Colon & 83.85 & 82.52 & \textbf{85.22} & 76.71 & 82.29 & 81.09 & 83.01 & 82.33 & 80.70 \\
Leukemia & 93.79 & \textbf{94.69} & 91.22 & 92.37 & 94.09 & 93.04 & 92.54 & 93.96 & 89.19 \\
Madelon & 60.13 & 58.38 & 60.04 & \textbf{63.14} & 61.96 & 58.48 & 61.07 & 61.04 & 55.74 \\
\hline
Mean & 74.64 & 74.56 & 73.07 & 73.37 & \textbf{77.22} & 75.52 & 76.86 & 78.62 & 73.76 \\
\hline
\end{tabular}
\end{table}

Against SVM-RFE, SALVES is ahead on Arcene by 2.5 points, tied on Prostate and Colon, and behind by 1.6 points on Leukemia and 0.9 points on Madelon. The two methods reach the same level in different ways. SVM-RFE trains a support vector machine on the dataset again at each step. SALVES applies a value model that was trained once, on synthetic tasks, without any real dataset. Selecting variables is itself useful on four of the five datasets, where SALVES is 2.3 to 10.9 points above the classifiers trained on all variables. Arcene is the exception, since its 10{,}000 variables together give a higher accuracy than every selection method.

The strength of SALVES in this application comes from the combination of the signals. When the eight association signals are averaged with equal weights, without any training, the accuracy is tied with that of SALVES on four datasets and is 7.3 points higher on Prostate, which puts it 1.8 points above SALVES on average. On Prostate, the best baseline is the logistic weights, the one association signal to which the value model gives a negative weight. The eight signals are therefore reliable whether they are combined with learned or with equal weights. This comparison is revisited in Section~\ref{sec:res-analyses}.

In summary, input selection shows that the value model alone, without search and without a diversity term, is enough for a reliable selection. Trained on synthetic tasks only, SALVES is tied with SVM-RFE, the best baseline, over the five real datasets and at every number of variables, and it is ahead of each of the five other baselines. \textcolor{black}{Unlike the four ranking baselines, it never fails on a dataset.} This reliability comes from combining multiple signals, and, in this application, equal weights perform at least as well as learned weights.
\subsection{Experimental design}\label{sec:res-design}
Experimental design differs from the three other applications in one respect: the value model does not score the candidates once, it guides each move of a search. A design is a table of experiments, in which each row is one experiment and each column one factor, set at one of three levels. A good design allows the effects of the factors to be estimated separately. The criterion defined in Section~\ref{sec:applications} quantifies the degree of correlation among these effects, with lower values indicating better designs. Simulated annealing improves a design by swaps, each of which exchanges two entries within one column.

The value model is a multilayer perceptron that predicts whether a swap will improve the criterion. Its training data cost nothing extra: during ordinary runs of simulated annealing, every swap that is tried is recorded together with its outcome. It is trained on 132{,}000 such swaps from designs of 11 sizes. The five test designs have other sizes, from 24 rows and 10 columns to 66 rows and 10 columns.

At each step, simulated annealing with random swaps draws one swap and computes the criterion of the resulting design. SALVES draws four swaps and computes the criterion only for the one with the highest predicted value. Both therefore use one evaluation of the criterion per step. Each receives 2{,}000 evaluations, and there are 50 runs per design, 250 in total. The intervals are computed over runs, with each design given the same weight.

With the value model, the search reaches a better design at the same budget. The criterion is 0.4878 on average, against 0.4899 with random swaps, a gain of 0.4\% whose interval excludes zero. SALVES is ahead on the two designs with 48 rows, by 1.7\% and 1.0\%, tied on the designs with 24 and 66 rows, and behind on the design with 25 rows and 12 columns, by 0.75\%. Random swaps need 10 to 13\% more evaluations to reach the same criterion.

The gain is smaller than in the other applications, which is consistent with the nature of the problem. In the other applications, the utility is a test accuracy that cannot be measured when the selection is made, and the value model supplies an estimate of it. In this application, the criterion can be computed exactly during the search. Simulated annealing therefore has access to the true utility of every swap it tries. The value model cannot supply information that the search lacks. It can only save evaluations. Section~\ref{sec:res-analyses} examines how the gain changes when the value model chooses among two or eight swaps instead of four.
\subsection{Analyses of the framework}\label{sec:res-analyses}
This subsection investigates which components of SALVES account for the results of Sections~\ref{sec:res-al} to~\ref{sec:res-design}. Each ablation modifies one component of SALVES and keeps the other settings unchanged, as described in Section~\ref{sec:ablations}. The first ablations examine the two components that define the framework, the value model and the diversity of the selected set. Their effects are reported for each application in Figures~\ref{fig:abl-al}, \ref{fig:abl-bands} and~\ref{fig:abl-inputs-design}. The remaining ablations investigate alternative value models, diversity formulations and search strategies, with results reported in Table~\ref{tab:abl-other}.
\subsubsection{Learning and diversity}
Removing the value model means replacing the predicted value, inside the same search, with the hand-crafted score traditionally used in each application: the classifier margin in active learning and the mutual information with the class in band selection. Removing the diversity means selecting the candidates with the highest predicted values. Unless a difference is reported as tied, its 95\% confidence interval excludes zero.

In active learning, both components contribute to the gain (Figure~\ref{fig:abl-al}). On the reduced pool of ImageNet, SALVES achieves an accuracy of 64.9\%. It outperforms the version without the value model by 1.0 points and the version without the diversity by 7.8 points. Without diversity, accuracy drops to 57.0\%, below the 58.8\% of random selection, because the images with the highest predicted values are highly similar to one another. The diversity in turn requires a good score: the same search with random scores is 5.5 points below SALVES. The gain therefore comes from the combination of the two components. Of the 1.76 points that separate SALVES from TypiClust, the strongest baseline, about 0.7 points are attributable to the search procedure with the margin as a score, and about 1.0 points come from the value model. Moreover, the value model performs as well as a perfect value estimator. With the true values, which require labels that are not available when the selection is made, accuracy is tied with that of SALVES. The digit pairs show the same pattern on a smaller scale. SALVES achieves 97.0\%. It outperforms the version without the value model by 0.7 points and the version without the diversity by 1.2 points. Neither of these two versions outperforms BADGE, the strongest baseline. Only the complete framework does.
\begin{figure}[!h]
\centering
\includegraphics[width=\textwidth]{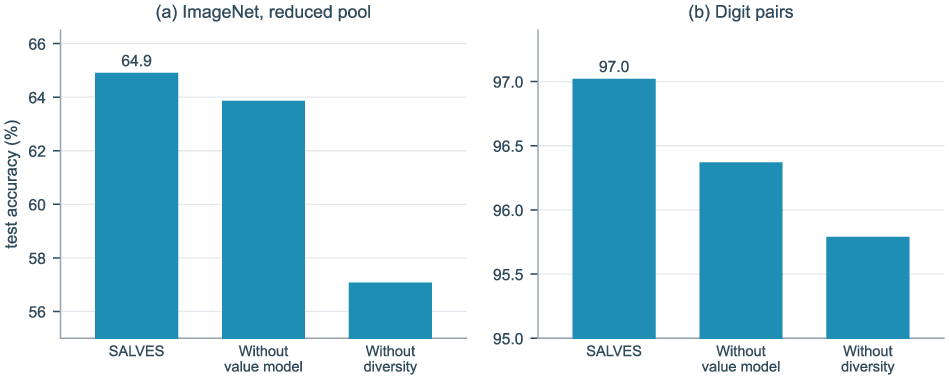}
\caption{Learning and diversity, active learning.}
\label{fig:abl-al}
\end{figure}

In band selection, SALVES achieves the highest accuracy of the three versions at budgets of three and ten bands (Figure~\ref{fig:abl-bands}). The relative contributions of the two components depend on the budget. With three bands, diversity contributes most: SALVES outperforms the version without diversity by 0.4 points, whereas replacing the value model with mutual information leaves the accuracy tied with that of SALVES. With ten bands, the roles are exchanged: SALVES outperforms the version without the value model by 0.3 points, and removing the diversity leaves the accuracy tied. Each component is also useful on its own. With three bands, either one adds approximately one point to the selection of the bands with the highest mutual information. With ten bands, either one adds 0.4 to 0.5 points. The versions with a diversity term use the same fixed weight in this figure, so that a single component is modified at a time. In Section~\ref{sec:res-bands}, SALVES also chooses this weight by cross-validation, which adds 0.7 points with three bands and 0.2 points with ten.
\begin{figure}[!h]
\centering
\includegraphics[width=\textwidth]{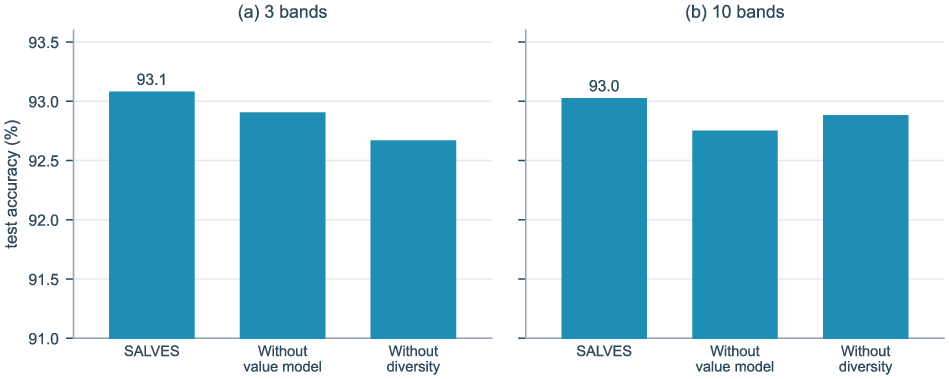}
\caption{Learning and diversity, band selection.}
\label{fig:abl-bands}
\end{figure}

Input selection and experimental design use no separate diversity term, so only the value model is changed (Figure~\ref{fig:abl-inputs-design}). In input selection, SALVES outperforms the version without the value model, the mutual information alone, by 2.2 points. This gain comes from combining multiple signals. The third bar keeps the eight association signals of Section~\ref{sec:applications} and replaces the learned weights by equal weights. The learned weights, trained on synthetic tasks only, are tied with equal weights on four of the five real datasets. On the fifth dataset, Prostate, equal weights are ahead, which places them 1.8 points above SALVES on average. With equal weights, the signals of SALVES also outperform SVM-RFE, the strongest baseline, by 1.4 points. In this application, the strength of SALVES therefore lies in its signals, and equal weights perform at least as well as learned weights.

In experimental design, the value model scores four random swaps at each step, and the search evaluates the one with the highest predicted value. Without the value model, the search evaluates one random swap. At 2{,}000 evaluations, SALVES improves the criterion from 0.490 to 0.488. The predicted values are informative: evaluating the swap with the lowest predicted value of the four worsens the criterion to 0.497. The number of swaps scored at each step sets the balance between the guidance of the value model and the randomness of the search. With two swaps, the criterion is tied with that of SALVES. With eight, it rises to 0.492, and the gain over random swaps is lost. A moderate guidance therefore serves the search best, and the four swaps of SALVES provide it.

\begin{figure}[!h]
\centering
\includegraphics[width=\textwidth]{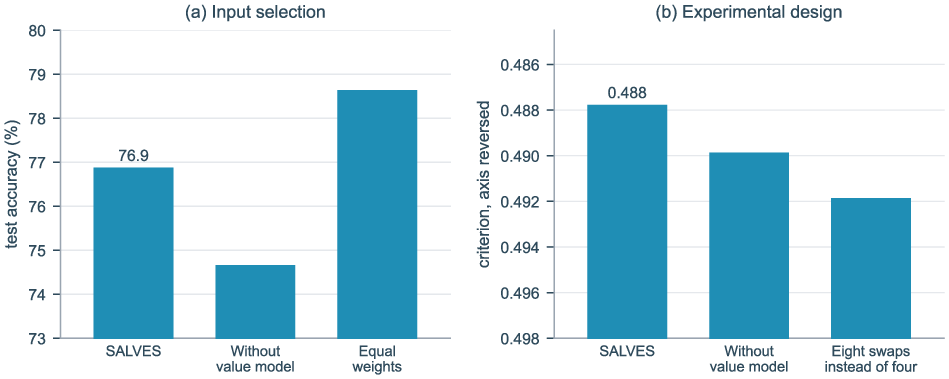}
\caption{Input selection and experimental design.}
\label{fig:abl-inputs-design}
\end{figure}

\subsubsection{Other ablations}
Table~\ref{tab:abl-other} reports the ablations that modify one component of SALVES without removing it. Each row gives the difference between SALVES and the modified version, with positive values indicating that SALVES is better. None of these modifications outperforms SALVES: every difference is either tied or in its favour.

\begin{table}[!h]
\centering
\small
\setlength{\tabcolsep}{4pt}
\caption{Other ablations of SALVES.}
\label{tab:abl-other}
\begin{tabular}{lrll}
\hline
Changed element & Difference & 95\% interval & Result \\
\hline
\multicolumn{4}{l}{Value model, reduced pool of ImageNet} \\
Random scores instead of predicted values & 5.52 & [5.40, 5.65] & Ahead \\
True values instead of predicted values & $-0.02$ & [$-0.09$, 0.05] & Tied \\
Set transformer & $-0.04$ & [$-0.16$, 0.07] & Tied \\
Graph network & $-0.09$ & [$-0.24$, 0.07] & Tied \\
Transformer & 0.00 & [$-0.11$, 0.12] & Tied \\
Convolutional network reading the image & $-0.01$ & [$-0.11$, 0.09] & Tied \\
38 signals instead of 7 & $-0.08$ & [$-0.17$, 0.01] & Tied \\
Training data divided by 600 & 0.17 & [0.07, 0.27] & Ahead \\
\hline
\multicolumn{4}{l}{Diversity} \\
ImageNet, 1 group per selected image instead of 1.5 & 0.48 & [0.39, 0.57] & Ahead \\
Digit pairs, R\'enyi entropy instead of the spread & 0.58 & [0.40, 0.75] & Ahead \\
\hline
\multicolumn{4}{l}{Search} \\
Digit pairs, greedy search & $-0.06$ & [$-0.14$, 0.02] & Tied \\
Band selection, greedy search & $-0.03$ & [$-0.09$, 0.03] & Tied \\
Band selection, exact solver & 0.01 & [$-0.04$, 0.05] & Tied \\
\hline
\end{tabular}
\end{table}

SALVES is robust to the type of the value model. On the reduced pool of ImageNet, a set transformer, a graph network, a transformer and a convolutional network that reads the image itself are all tied with the multilayer perceptron of SALVES, as is a value model that reads 38 signals instead of seven. On the full pool, up to 50 labels per class, the four models remain within 0.3 points of it, the two transformers being slightly ahead. On the digit pairs, seven other models, from kernel models and gradient boosting to larger perceptrons and a set transformer, are tied with it, and none is ahead. A small multilayer perceptron with seven signals is therefore sufficient. The value model also requires little training data. Fitted on 600 times less data, nine states with 100 candidates each instead of 54 states with 10{,}000, SALVES loses only 0.2 points and remains 1.6 points above TypiClust. The learning phase is therefore inexpensive. In input selection, the simplest model transfers best. A multilayer perceptron trained on the same synthetic tasks as the logistic regression of Section~\ref{sec:applications} is 10.6 points below it on average, although it was slightly better on the synthetic validation tasks. A model that more closely fits the synthetic tasks may therefore generalise less effectively to real data.
A model that fits the synthetic tasks more closely can therefore transfer less well to real data.

The form of the diversity chosen for SALVES is confirmed on the test tasks. On the reduced pool of ImageNet, the group constraint uses 1.5 groups per selected image. With one group per selected image, the accuracy is 0.5 points lower. On the digit pairs, the diversity term of SALVES, the spread of Section~\ref{sec:applications}, outperforms the R\'enyi entropy by 0.6 points.

Simulated annealing achieves the accuracy of an exact solver at a small fraction of its cost. In band selection, the exact solver is tied with it. This solver proves the optimum in nine runs out of ten within its limit of ten minutes and takes from two seconds to several minutes per selection, depending on the weight of the diversity term, whereas simulated annealing takes 0.05 seconds. A greedy search, which adds one candidate at a time, is also tied with simulated annealing, on the digit pairs and in band selection. The quality of the selection is therefore determined by the objective, that is, by the value model and the diversity, rather than by the search that optimises it.

In summary, the gain of SALVES comes from combining a learned value with diversity. In active learning, both components contribute, and the complete framework achieves the largest gain over the best baseline. In band selection, SALVES achieves the highest accuracy of the three versions, the diversity contributing most with three bands and the value model with ten. In input selection, the gain comes from combining multiple signals, with learned or with equal weights. In experimental design, a moderate guidance by the value model improves the search. The framework is also robust. On ImageNet, a small value model with seven signals and little training data is sufficient. In band selection, simulated annealing achieves the accuracy of an exact solver. None of the changes tested in Table~\ref{tab:abl-other} outperforms SALVES.
\section{Conclusion}
The central finding of this paper is that the value of a candidate does not have to be designed by hand for each selection problem. It can be learned once, on tasks in which the hidden information is known, and then reused on new tasks. Used inside a simple selection step, this learned value is sufficient to match or outperform the rules designed for batch active learning, band selection, input selection and experimental design.

Three observations stand out. First, the benefit of a learned value is largest at small selection budgets. Second, the value model remains effective on tasks that differ from its training tasks, including real data after training on synthetic tasks, which indicates that the signals keep their meaning from one task to another. Third, in active learning and band selection, the gain comes from the value model and from diversity together, and not from the choice of the search algorithm.

SALVES applies when training tasks with known hidden information are available, real or synthetic, and when no single hand-crafted score is reliable across tasks and budgets. When a classical rule is already close to the best choice, as margin sampling is once many labels are available, that rule leaves little room for improvement.
}
\bibliographystyle{unsrtnat}
\bibliography{references}
\end{document}